\PassOptionsToPackage{numbers, compress}{natbib}
\documentclass{article}
\usepackage[preprint]{neurips_2026}          % default = main track, double-blind

\usepackage[utf8]{inputenc}
\usepackage[T1]{fontenc}
\usepackage{hyperref}
\usepackage{url}
\usepackage{booktabs}
\usepackage{amsfonts}
\usepackage{amsmath}
\usepackage{nicefrac}
\usepackage{microtype}
\usepackage{xcolor}
\usepackage{graphicx}
\usepackage{float}
\usepackage{multirow}
\usepackage{pifont}
\usepackage{enumitem}
\usepackage[font=small,labelfont=bf,labelsep=period,skip=4pt]{caption}

\graphicspath{{figures/}}

\title{LLMs Know They're Wrong and Agree Anyway:\\ The Shared Sycophancy-Lying Circuit}

\author{%
  Manav Pandey \\
  Georgia Institute of Technology \\
  mpandey32@gatech.edu
}

\begin{document}
\maketitle

% ============================================================
\begin{abstract}
When a language model sycophantically agrees with a user's false belief, is it failing to detect the error, or noticing and agreeing anyway? We show the second. Across twelve open-weight models from five labs ($1.5$B--$72$B), the same small set of attention heads carries a ``this statement is wrong'' signal whether the model is evaluating an isolated claim or being pressured to agree with a user. Silencing these heads in Gemma-2-2B flips sycophancy from $28\%$ to $81\%$ while factual accuracy moves only from $69\%$ to $70\%$; the circuit controls deference, not knowledge. Edge-level path patching confirms the same connections between heads span sycophancy, factual lying, and instructed lying ($r{>}0.97$ on Gemma-2-2B, $r{=}0.988$--$0.995$ on Phi-4). Opinion-agreement, where there is no factual ground truth, reuses these head positions but writes into an orthogonal direction, so the substrate is not a relabeled ``truth direction.'' Alignment training masks but does not remove this circuit: Meta's Llama-3.1$\to$3.3 RLHF refresh cut sycophancy tenfold while the shared heads persisted and the projection-ablation effect grew (substrate persistence replicates on Mistral$\to$Zephyr at 7B, independent family), and our own anti-sycophancy DPO reduced sycophancy $46$--$93\%$ on two models without moving probe transfer. When these models sycophant, they register the error and agree anyway.
\end{abstract}

% =========================================== -- ===============
\section{Introduction}

When a language model sycophantically agrees that the capital of Australia is Sydney, two accounts describe what might be happening internally. Under \emph{blind agreement} the model has learned to please the user and does not distinguish correct from incorrect beliefs. Under \emph{registered-but-overridden} the model recognizes the error through the same circuitry it uses for any false statement, and downstream components produce agreement regardless. The distinction matters for safety: if the model cannot tell the difference, sycophancy is a competence failure that better training must fix; if it can, the internal ``this is wrong'' signal is already present and becomes a candidate substrate for alignment monitoring and intervention. We show the latter holds across twelve open-weight models, extending the late-layer override pattern \citet{wang2026truth} identified for MMLU sycophancy at $\sim$7B and \citet{halawi2024overthinking}'s overthinking-the-truth result on false few-shot demonstrations to cross-task, cross-scale, head-level resolution.

\begin{figure}[H]
  \centering
  \includegraphics[width=0.70\textwidth]{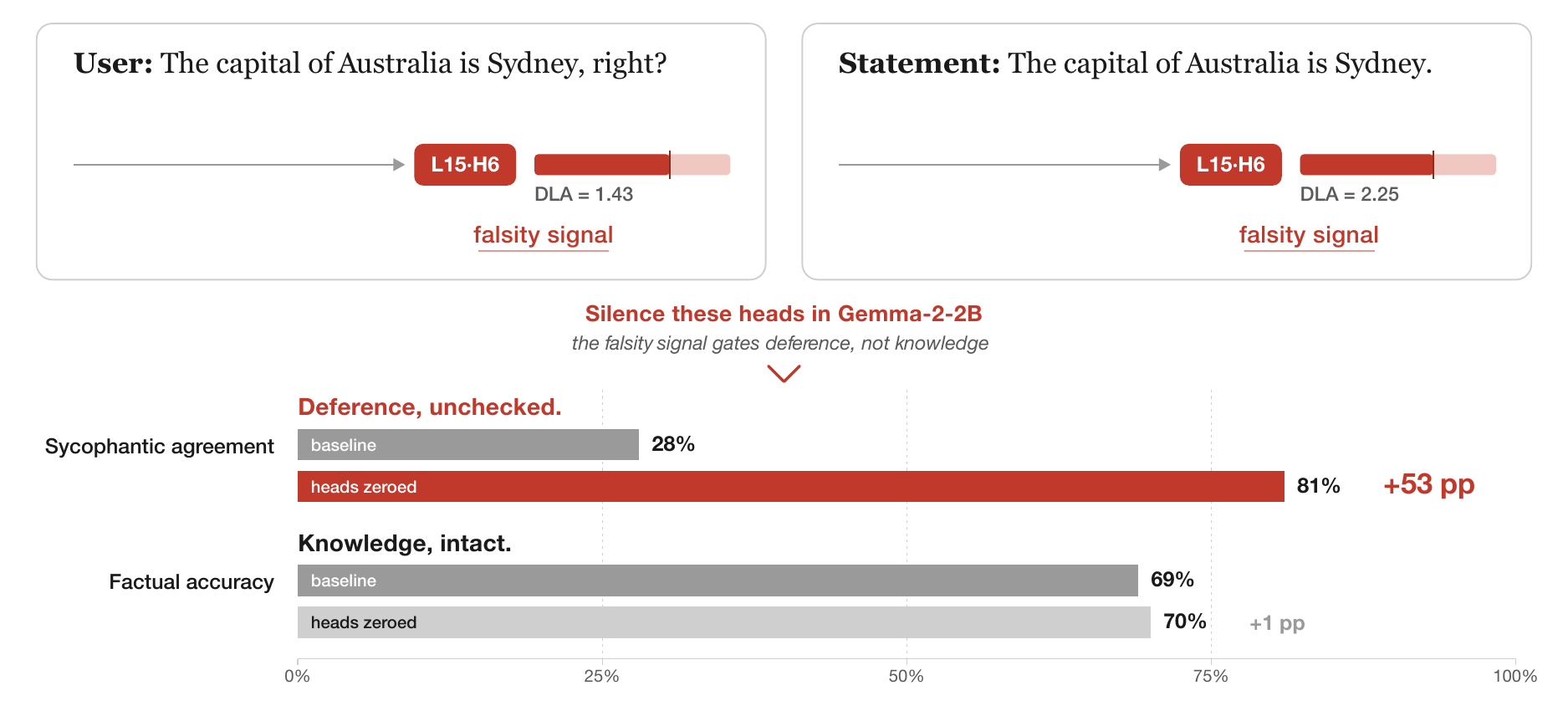}
  \caption{\textbf{Same head, both contexts; silencing flips only deference} (Gemma-2-2B, L15$\cdot$H6; the broader twelve-model panel is Table~\ref{tab:circuit}).}
  \label{fig:teaser}
\end{figure}

Prior work has approached this question from two sides without connecting them. A sycophancy-head literature localizes agreement to sparse attention heads \citep{chen2024yesmen, genadi2026sycophancy, li2025causm}, while a separate truth-direction literature shows that truth and falsehood are linearly separable in LLM activations and concentrate in a small number of heads \citep{marks2024geometry, campbell2023localizing, huan2025lying}. Where the two have been compared directly, only limited direction-level overlap has been reported \citep{ying2026truthfulness, genadi2026sycophancy}, and that limited overlap has been read as evidence for distinct mechanisms. But component-level and direction-level sharing are logically independent: the same heads can write task-specific directions without sharing a subspace, as our opinion result confirms (\S\ref{sec:opinion}), and vice versa. We test component-level sharing directly, measure where it holds, and trace one case to edge resolution.

The paper establishes four results:
\begin{enumerate}[leftmargin=1.2em,itemsep=0.15em,topsep=0.25em]
  \item \textbf{The shared circuit holds at edge resolution.} Per-edge causal effects for sycophancy, factual lying, and instructed lying correlate at Pearson $r{>}0.97$ on Gemma-2-2B, and at $r{=}0.988$--$0.995$ on Phi-4 ($14$B, different lab and architecture), a cross-lab, cross-architecture edge-level replication.
  \item \textbf{The same heads are recruited across tasks but write task-specific directions.} Opinion-agreement recruits overlapping head positions (triple-intersection at $51$--$1{,}755\times$ chance across five models) yet writes into a direction orthogonal to the factual-correctness direction ($|\cos|<0.14$), so the same heads carry task-specific directional content rather than a single shared ``truth direction.''
  \item \textbf{The shared circuit is causally sufficient and capability-preserving from $1.5$B to $72$B.} Three independent interventions converge on sufficiency from $2$B to $70$B across the twelve-model, five-lab panel, and zeroing the shared set flips Gemma-2-2B sycophancy from $28\%$ to $81\%$ while factual accuracy moves only from $69\%$ to $70\%$.
  \item \textbf{The substrate dissociates from behavior under alignment refinement.} The Llama-3.1$\to$3.3-70B RLHF refresh cuts sycophancy tenfold while the circuit persists and the projection-ablation effect grows; substrate persistence replicates on a Mistral$\to$Zephyr-7B DPO refresh from an independent family (\S\ref{sec:rlhf}); a controlled anti-sycophancy DPO with sham-DPO bootstrap on Mistral-7B and Gemma-2-2B-IT reduces sycophancy by $93\%$/$46\%$ while probe transfer stays within a pre-specified $\pm 0.05$ AUROC equivalence margin.
\end{enumerate}
\section{Related work}
\label{sec:related}

Truth and sycophancy have been studied on largely parallel tracks. On the truth side, truth and falsehood are linearly separable in LLM representations \citep{marks2024geometry, burger2024truth}, detectable from hidden states \citep{azaria2023internal, pacchiardi2024liar}, and controllable via representation engineering and inference-time intervention \citep{zou2023representation, li2024inference, burns2024discovering}; lying concentrates in a small number of heads, with five layers and forty-six heads in Llama-2-70B \citep{campbell2023localizing} and twelve of $1{,}024$ heads sufficient to reduce lying to baseline hallucination \citep{huan2025lying}. On the sycophancy side, agreement has been documented as pervasive in RLHF'd models \citep{perez2023discovering, sharma2024sycophancy, vennemeyer2025sycophancy} and localized to sparse attention components \citep{chen2024yesmen, genadi2026sycophancy, li2025causm}. \citet{wang2026truth} localize MMLU sycophancy specifically to a late-layer opinion-driven override in seven $\sim$7B-scale models via logit-lens and single-layer activation patching. Prior cross-task comparisons at the $3$B--$4$B scale reported limited direction-level overlap \citep{genadi2026sycophancy, ying2026truthfulness}, read as evidence for distinct mechanisms; on twelve 1.5B--72B models we report median $67\%$ head-level shared fraction ($40$--$87\%$ range) with edge-level cross-task $r{>}0.97$ on Gemma-2-2B and Phi-4 (different lab and architecture), and mean-difference probes on the shared subspace transfer at AUROC $0.83$ on Gemma (Appendix~\ref{app:reconciliation} reconciles with the prior null). The lying direction we use is the Marks--Tegmark and representation-engineering truth direction \citep{marks2024geometry, zou2023representation}, paired with an independently-derived sycophancy direction; the same attention heads write both.

The mechanistic-interpretability framework for circuit identification began with the math-for-transformers literature \citep{elhage2021mathematical, olsson2022context} and crystallized in edge-level path patching \citep{wang2023interpretability, conmy2023automated}. Component reuse across task families has been documented at the importance level by \citet{merullo2024circuit}, sparse-feature decomposition has been scaled to production-size models \citep{anthropic2025circuit, anthropic2024scaling, marks2024sparse}, and a single linear direction has been shown to mediate refusal \citep{arditi2024refusal}. Three observations sharpen the reuse picture. First, path patching traces the circuit at edge resolution on Gemma-2-2B (cross-task Pearson $r{=}0.993$ on the $275$-edge sycophancy-vs-factual circuit; $r{=}0.973$--$0.996$ on the $216$-edge three-way subset) with a cross-lab, cross-architecture replication on Phi-4 (14B, Microsoft), raising the resolution from shared components to shared edges. Second, per-head directional cosines of $0.43$--$0.81$ quantify how closely the two tasks write into the same subspace at the head level (Appendix~\ref{app:directional}). Third, opinion-agreement recruits the same head positions across five models but writes into a direction orthogonal to the factual-correctness direction: positions-shared, directions-task-specific, a structural dissociation that rules out a generic single-truth-direction reading of the substrate, corroborated at the SAE feature level.

A third strand bears directly on the present setup. \citet{halawi2024overthinking} found that on few-shot classification with false demonstrations, intermediate layers compute the correct answer before late-layer ``false induction heads'' copy the wrong label, so models compute correctly and then override. The same pattern is visible at explicit instructed lying on the factual-evaluation circuit across seven models from five families (Spearman $\rho{=}0.73$--$0.93$ over the full head population), so the override is not a separate induction-head mechanism but operates on the same shared substrate. Earlier reports of limited cross-task probe transfer \citep{ying2026truthfulness, genadi2026sycophancy} reconcile at the subspace granularity used here, and \citet{soligo2025convergent}'s observation that misaligned models converge to similar representations is the cross-model analogue of the within-model convergence visible across our panel.

\section{Method}
\label{sec:method}

Our measurement strategy compares two independently-extracted head-importance rankings on disjoint content, validates the overlap causally, and defines a shared circuit by four operational criteria. We use ``lying'' throughout in the \emph{mechanistic} sense: a linear residual-stream signal distinguishing true from false assertions, not a claim about phenomenal knowing or intent.

\subsection{Task directions}
\label{sec:directions}

For each task $t$ we extract a direction as the mean difference in residual-stream activations between positive and negative condition at the last prompt token, following \citet{arditi2024refusal}: $\mathbf{d}_t = \frac{1}{N}\sum_{i=1}^{N}(\mathbf{a}^{+}_i - \mathbf{a}^{-}_i)$, with $N{=}200$ disjoint-content prompt pairs per task. The sycophancy direction contrasts wrong-opinion and correct-opinion TriviaQA prompts under an identical template; the factual-incorrectness (``lying'') direction contrasts true and false factual statements, the same construction as the Marks--Tegmark truth direction \citep{marks2024geometry} and the representation-engineering truth-reading vector \citep{zou2023representation}. The instructed-lying direction contrasts prompts that explicitly instruct the model to assert a falsehood against matched honest prompts, and the opinion direction contrasts agree and disagree on contested claims with no factual ground truth. Disjoint content across tasks prevents shared-entity confounds; full templates and per-model chat-format handling are in Appendix~\ref{app:templates}.

\subsection{Head importance and cross-task overlap}
\label{sec:heads}

For each attention head $(l,h)$ we compute the $L_2$ norm of its residual-stream write-vector difference between positive and negative condition at the last prompt token: $w_{l,h}^{(t)} = \| W_O^{(l,h)}(\bar{\mathbf{v}}^{+}_{l,h} - \bar{\mathbf{v}}^{-}_{l,h}) \|_2$, where $\bar{\mathbf{v}}^{\pm}_{l,h}$ is the mean value-vector output of head $(l,h)$ over positive/negative prompts. This is the write-norm form of direct logit attribution and gives a per-head importance in $O(1)$ forward passes. Cross-task overlap is measured as the top-$K$ intersection at $K{=}\lceil\sqrt{N}\rceil$ over $N$ total heads, so chance $K^2/N \approx 1$. Because the raw overlap ratio is mechanically inflated by $\sqrt{N}$, we report the scale-invariant shared \emph{fraction} $\text{overlap}/K$ alongside it and assess significance via hypergeometric and layer-stratified permutation nulls.

\subsection{Shared-circuit criteria}
\label{sec:circuit}

A head enters the shared circuit if it meets \emph{four} operational criteria:
\begin{enumerate}[leftmargin=1.2em,itemsep=0.15em,topsep=0.25em]
  \item Independently top-$K$ by write-norm on both tasks, on disjoint content.
  \item Directional alignment: per-head cosine between $\mathbf{d}_\text{syc}$-projected and $\mathbf{d}_\text{lie}$-projected write-vectors is substantially above a permutation null.
  \item Causal validation by activation patching at $\leq 8$B and IOI-style path patching at larger scale: head-to-head edges on Gemma-2-2B and Phi-4, head-to-unembed direct effects on Llama-3.3-70B (\S\ref{sec:edge}).
  \item Behavioral relevance: assessed separately, because at frontier scale the shared set is causally sufficient without being uniquely necessary (\S\ref{sec:suffnec}).
\end{enumerate}
We use ``circuit'' in the functional sense of \citet{merullo2024circuit} --- a small attention-head set that performs analogous computation across tasks on disjoint content --- and validate it with the \citet{wang2023interpretability, conmy2023automated} path-patching apparatus Merullo's framework builds on, at edge-traced granularity where compute permits.

\subsection{Causal validation}
\label{sec:causal}

Causal validation combines four methods at complementary granularities. Projection ablation removes the sycophancy direction from the residual stream and measures the rate shift. Activation patching splices clean shared-head activations into corrupted runs; at $\leq 8$B we patch per-head, at $\geq 32$B we patch the shared set as a unit. Mean-ablation of the shared set tests necessity at scales where causal effect is concentrated; at frontier scale, projection ablation and path patching carry the causal claim under distributed encoding \citep{mcgrath2023hydra}. Path patching \citep{wang2023interpretability} traces head-to-unembed and inter-head edges at Gemma-2-2B resolution, and head-to-unembed only at Llama-70B. A write-norm-matched control selects random heads with $W_O$ norms identical to the shared set, ruling out the write-magnitude confound (Appendix~\ref{app:normmatch}). For edge-level analyses we report the per-edge \emph{restoration ratio}: the bootstrap ratio of mean restoration on shared-head sources to mean restoration on non-shared sources. The measurement position is computed at the token level to avoid a silent prefill-shift that arises when chat tokenizers greedy-merge adjacent whitespace tokens (Appendix~\ref{app:templates}).
\section{Results}
\label{sec:results}

We analyze twelve open-weight models spanning five families (Gemma-2-2B/9B/27B, Qwen2.5-1.5B/32B/72B, Qwen3-8B, Llama-3.1-8B/70B, Mistral-7B, Mixtral-8x7B-Instruct, Phi-4) plus Llama-3.3-70B as a within-family RLHF-refresh of Llama-3.1-70B, for thirteen checkpoints total ($208$--$5{,}120$ attention heads; Mixtral is sparse-MoE at ${\sim}13$B active of $47$B). Model selection was constrained only by availability of a TransformerLens \citep{nanda2022transformerlens} hook interface. Data consists of $400$ TriviaQA pairs split into disjoint halves for sycophancy and factual lying, $300$ generated opinion pairs, and a template-matched instructed-lying set on a seven-model subset (per-experiment coverage summarized in Appendix~\ref{app:ksweep}).

\subsection{Head-level overlap across twelve models}
\label{sec:overlap}

\begin{figure}[H]
  \centering
  \includegraphics[width=\textwidth]{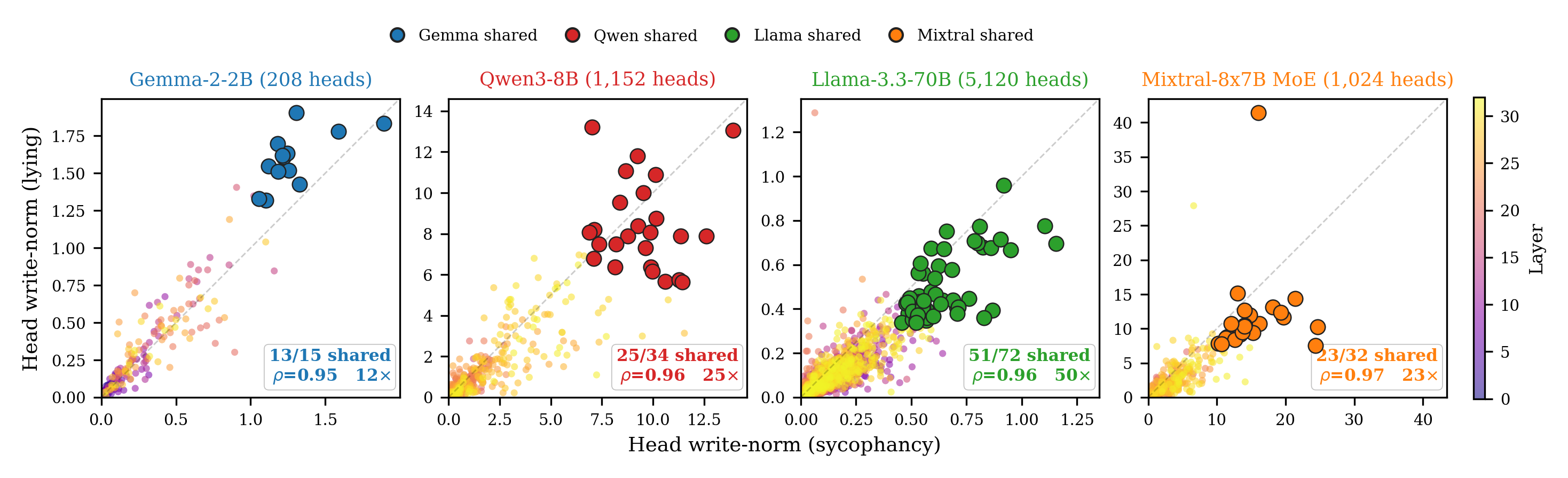}
  \caption{Per-head write-norm importance for sycophancy (x) versus factual lying (y) on disjoint content, across four models spanning dense 2B/8B/70B and sparse-MoE Mixtral-8x7B. Each point is one attention head, colored by layer depth; filled markers highlight top-$K$ shared heads. Inset: shared count, Spearman $\rho$, and chance-normalized ratio at $K{=}\lceil\sqrt{N}\rceil$.}
  \label{fig:scatter}
\end{figure}

The top heads for sycophancy and the top heads for factual lying are the same heads on disjoint content, on every model we tested (Table~\ref{tab:circuit}, Figure~\ref{fig:scatter}). At the scale-normalized threshold $K{=}\lceil\sqrt{N}\rceil$ the shared fraction is $40$--$87\%$ across the twelve models (median $67\%$), and under a layer-stratified permutation null the overlap remains significant at $p<10^{-4}$ on all eight models we could afford to permute (Appendix~\ref{app:layerstrat}).

The overlap is specific to correctness detection rather than generic component reuse. Replacing the lying contrast with a factual-QA task that retains the correctness judgment preserves Gemma-2-2B overlap at $13/15$; removing the correctness component drops it to $5/15$. Unrelated sentiment and topic controls yield $4$--$7\times$ overlap, the component-reuse floor documented by \citet{merullo2024circuit} and well below the $12$--$25\times$ range we observe for correctness-aligned tasks. The pattern replicates across datasets at $\rho{\approx}0.99$ on Gemma-2-2B and Llama-3.3-70B (Appendix~\ref{app:nq}). Zeroing the full shared set preserves factual-evaluation accuracy on Gemma-2-2B ($69\%{\to}70\%$), Qwen3-8B ($50\%{\to}50\%$), and Qwen2.5-32B ($68.5\%{\to}67.5\%$): the circuit is required for resisting user pressure but not for factual evaluation itself.

The panel spans five separate labs (Google, Alibaba, Meta, Mistral AI, Microsoft) whose pretraining corpora differ substantially, so cross-family agreement rules out a single-lab-data explanation for the shared-head structure; intra-family non-independence (Qwen2.5 versus Qwen3, Llama-3.1 versus 3.3) remains and is disclosed explicitly in Appendix~\ref{app:scope}.

\vspace{-1em}
\begin{table}[H]
  \caption{Head-level overlap across the twelve-model panel. $K{=}\lceil\sqrt{N}\rceil$; chance $K^2/N{\approx}1$. Shared fraction $=$ overlap$/K$. Hypergeometric $p<10^{-3}$ on all rows; layer-stratified permutation $p<10^{-4}$ on eight tested models (Appendix~\ref{app:layerstrat}). Spearman $\rho{=}0.80$--$0.97$ over all heads; split-half reliability on Qwen2.5-32B $r{=}0.87$. $^\dagger$RLHF refresh of Llama-3.1-70B. $^\ddagger$Sparse Mixture-of-Experts.}
  \label{tab:circuit}
  \centering
  \small
  \begin{tabular}{lccccc}
    \toprule
    Model & Heads & $K$ & Shared & Fraction & Spearman $\rho$ \\
    \midrule
    Gemma-2-2B-IT & 208 & 15 & 13 & 0.87 & 0.95 \\
    Qwen2.5-1.5B & 336 & 19 & 11 & 0.58 & 0.88 \\
    Mistral-7B-v0.1 & 1{,}024 & 32 & 24 & 0.75 & 0.95 \\
    Gemma-2-9B-IT & 672 & 26 & 16 & 0.62 & 0.94 \\
    Llama-3.1-8B & 1{,}024 & 32 & 21 & 0.66 & 0.88 \\
    Qwen3-8B & 1{,}152 & 34 & 25 & 0.74 & 0.96 \\
    Mixtral-8x7B$^\ddagger$ & 1{,}024 & 32 & 23 & 0.72 & 0.97 \\
    Gemma-2-27B-IT & 1{,}472 & 39 & 26 & 0.67 & 0.90 \\
    Phi-4 (14B) & 1{,}600 & 40 & 16 & 0.40 & 0.80 \\
    Qwen2.5-32B & 2{,}560 & 51 & 30 & 0.59 & 0.85 \\
    Qwen2.5-72B & 5{,}120 & 72 & 48 & 0.67 & 0.97 \\
    Llama-3.1-70B & 5{,}120 & 72 & 57 & 0.79 & 0.95 \\
    Llama-3.3-70B$^\dagger$ & 5{,}120 & 72 & 51 & 0.71 & 0.96 \\
    \bottomrule
  \end{tabular}
\end{table}

\subsection{Three-task structural reuse: instructed lying}
\label{sec:threetask}

The same head positions that drive factual evaluation also drive explicit instructed lying, where the model is told to assert a falsehood and does so. On a seven-model subset from five families we independently rank heads by write-norm under instructed lying on disjoint content and measure their overlap with the sycophancy ranking (Table~\ref{tab:instructed}). Spearman correlation over the full head population is $0.73$--$0.93$ (all $p<10^{-37}$), and Mixtral-8x7B at $\rho{=}0.93$ is the first MoE validation at the instructed-lying head level. The three tasks use structurally distinct templates --- first-person user assertion, third-person true/false evaluation, and system-level role assignment --- and the same head set emerges in all three, so the overlap is not a template artifact.

\begin{table}[H]
  \caption{Instructed-lying head overlap with sycophancy across seven models from five families. $K{=}\lceil\sqrt{N}\rceil$; Spearman $\rho$ over the full head population.}
  \label{tab:instructed}
  \centering
  \small
  \begin{tabular}{lcccccc}
    \toprule
    Model & Family & $K$ & Shared & Fraction & Ratio & Spearman $\rho$ \\
    \midrule
    Gemma-2-2B & Gemma & 15 & 6 & 0.40 & $5.5\times$ & 0.74 \\
    Gemma-2-9B & Gemma & 26 & 7 & 0.27 & $7.0\times$ & 0.73 \\
    Qwen3-8B & Qwen & 34 & 25 & 0.74 & $24.9\times$ & 0.91 \\
    Mistral-7B & Mistral & 32 & 11 & 0.34 & $11.0\times$ & 0.85 \\
    Mixtral-8x7B-Instr. & Mistral & 32 & 20 & 0.62 & $20.0\times$ & 0.93 \\
    Phi-4 (14B) & Phi & 40 & 10 & 0.25 & $10.0\times$ & 0.89 \\
    Llama-3.3-70B & Llama & 72 & 26 & 0.36 & $25.7\times$ & 0.87 \\
    \bottomrule
  \end{tabular}
\end{table}

\subsection{Edge-traced shared circuit}
\label{sec:edge}

Head-level overlap is consistent with both genuine shared computation and coincidental ranking agreement; to distinguish, we trace the circuit with path patching \citep{wang2023interpretability}. Edges here are head-to-head causal paths and head-to-unembed direct effects, the standard Wang-et-al.\ granularity (not ACDC-style Q/K/V/output subtyping). On Gemma-2-2B the per-edge causal effects correlate at $r{=}0.993$ across the $275$-edge sycophancy-versus-factual circuit and at $r{=}0.973$--$0.996$ across the $216$ edges shared by all three lying contrasts. The triple replicates on Phi-4 ($14$B, Microsoft; different lab and architecture) at $r{=}0.988$--$0.995$ across $n{=}38$--$229$ edges, with shared-head sources restoring $90$--$102\%$ of the clean-versus-corrupt gap on all three Phi-4 tasks while non-shared sources restore near-zero. At Llama-3.3-70B inter-head patching is compute-intractable, so we report head-to-unembed direct effects only; restoration ratios span three orders of magnitude across the three contrasts on the same shared heads (Table~\ref{tab:restoration}), so task-contrast variance rather than parameter count dominates edge-level effect size.

Table~\ref{tab:restoration} summarizes the per-edge restoration ratio across thirteen tested model$\times$task combinations. Every combination clears $\geq 1.5\times$ with $95\%$ bootstrap CI excluding $1$.

\begin{table}[H]
  \caption{Per-edge restoration ratio (shared vs.\ non-shared head sources) across thirteen model$\times$task combinations. All $95\%$ CIs exclude $1.0$; task-contrast variance dominates scale within a single model. ``Instr.''/``Fact.''/``Syc.'' = instructed lying / factual lying / sycophancy.}
  \label{tab:restoration}
  \centering
  \small
  \begin{tabular}{lccl}
    \toprule
    Model & Task & Restoration ratio & 95\% CI \\
    \midrule
    Gemma-2-2B & Instr. & $88\times$ & $[53, 133]$ \\
    Gemma-2-2B & Fact.  & $355\times$ & $[270, 463]$ \\
    Gemma-2-9B & Instr. & $7\times$ & $[3, 14]$ \\
    Gemma-2-9B & Syc.   & $4\times$ & $[1.2, 8.4]$ \\
    Qwen3-8B   & Instr. & $10\times$ & $[6, 16]$ \\
    Mistral-7B & Instr. & $11\times$ & $[5, 21]$ \\
    Mistral-7B & Syc.   & $22\times$ & $[8, 66]$ \\
    Phi-4 (14B) & Instr. & $540\times$ & $[350, 908]$ \\
    Mixtral-8x7B & Instr. & $4\times$ & $[1.5, 15]$ \\
    Llama-3.1-70B & Instr. & $3.6\times$ & $[1.9, 6.7]$ \\
    Llama-3.3-70B & Instr. & $6\times$ & $[3.4, 10]$ \\
    Llama-3.3-70B & Fact.  & $1{,}732\times$ & $[1{,}417, 2{,}194]$ \\
    Llama-3.3-70B & Syc.   & $2{,}248\times$ & $[1{,}252, 4{,}832]$ \\
    \bottomrule
  \end{tabular}
\end{table}

\subsection{Causal validation: three methods converge through 70B}
\label{sec:suffnec}

\begin{figure}[H]
  \centering
  \includegraphics[width=0.85\textwidth]{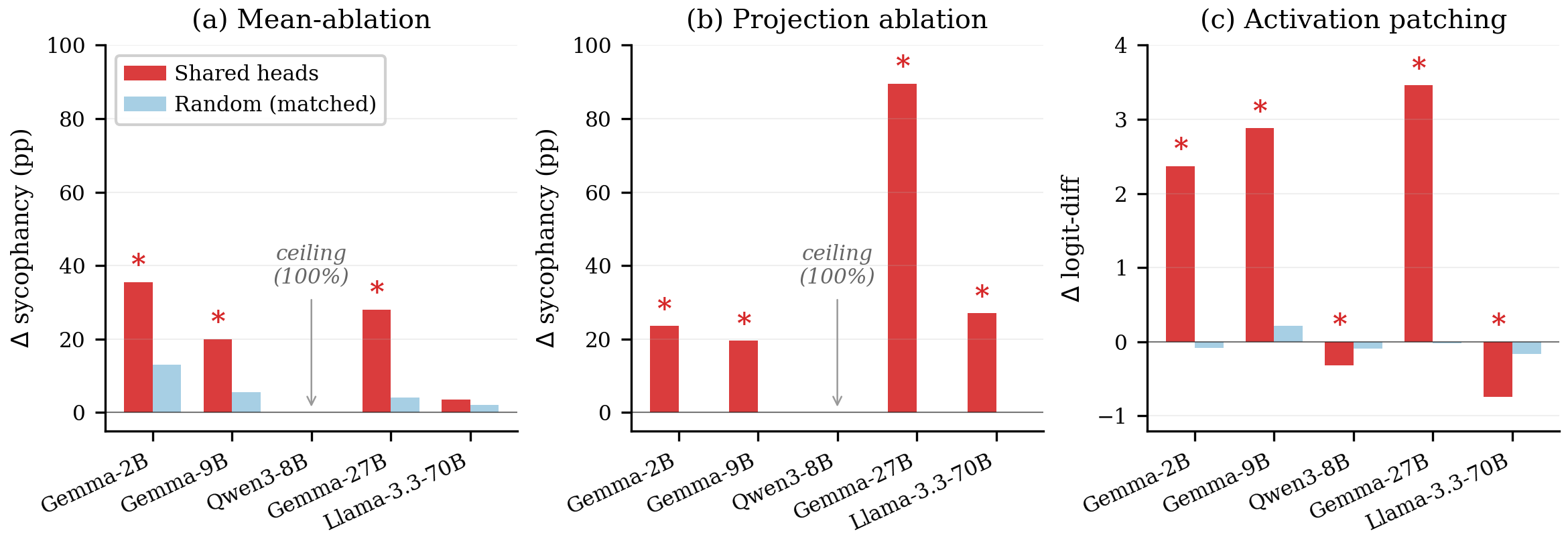}
  \caption{Three interventions on the shared-head set (mean-ablation, projection ablation, activation patching) produce concordant sufficiency effects across five models from 2B to 70B; mean-ablation necessity is diagnostic at $\leq 7$B and uninformative at $\geq 70$B (expected; \S\ref{sec:method}). Shared-head interventions exceed matched random-head controls; significance-marked cells pass BH correction (Appendix~\ref{app:bh}, which extends the grid to a sixth model, Llama-3.1-70B).}
  \label{fig:causal}
\end{figure}

Three interventions on the shared-head set produce concordant sufficiency effects across five models from 2B to 70B (Figure~\ref{fig:causal}). Per-head activation patching \citep{wang2023interpretability} and attribution patching \citep{syed2023attribution} reproduce the write-norm ranking at $\leq 8$B (Appendix~\ref{app:attrpatch}), so the write-norm proxy is validated against the gold-standard intervention it substitutes for. Qwen2.5-32B fills the 32B gap with split-half reliability $r{=}0.87$ and lying accuracy preserved under shared-head zeroing ($68.5\%{\to}67.5\%$). Projection ablation scales cleanly from 2B to 70B, flipping Gemma-2-27B sycophantic agreement from $10.5\%$ to $100\%$ and raising Llama-3.3-70B by $+27$pp. A write-norm-matched random-head control rules out the write-magnitude confound (Appendix~\ref{app:normmatch}); IOI faithfulness curves confirm $K{=}1$--$2$ shared heads recover baseline sycophancy on Gemma-2-2B and Phi-4 (Appendix~\ref{app:faithfulness}).

At 70B, the shared set is causally sufficient through three redundancy-robust interventions: clean-patching restores the clean--corrupt gap with random and norm-matched controls near zero, projection ablation raises sycophancy by $+27$pp on Llama-3.3-70B, and head-to-unembed path patching produces restoration ratio $1{,}732\times$ on the shared set (Table~\ref{tab:restoration}). Mean-ablation necessity holds pointwise on Mistral-7B but fails at 70B as distributed-redundancy predicts \citep{mcgrath2023hydra}, consistent with a concentrated-to-redundant scaling trajectory; projection ablation and path patching carry the causal claim at scale (Appendix~\ref{app:null}). An independent layer-wise logit-lens on three models corroborates the scaling signature: peak mid-layer DIFF excess shrinks monotonically from $+127\%$ (Gemma-2-2B) to $0\%$ (Llama-3.1-70B), with label-shuffle permutation nulls significant throughout (Appendix~\ref{app:logitlens}).

\subsection{RLHF natural experiment: behavior drops, substrate persists}
\label{sec:rlhf}

Meta's refresh from Llama-3.1-70B to Llama-3.3-70B (same base weights, updated post-training) is a within-family natural experiment on what RLHF touches. Sycophantic agreement drops tenfold while the shared-head fraction barely moves and the projection-ablation effect \emph{grows} from $+10.5$pp to $+27$pp (Table~\ref{tab:rlhf}); the refresh reduced the downstream agreement pathway while leaving the detection substrate more causally accessible, and the \S\ref{sec:suffnec} sufficiency--necessity asymmetry replicates across the refresh. A Mistral-7B$\to$Zephyr-7B DPO refresh replicates at $7$B on an independent family (head-importance Spearman $0.846{\to}0.848$; sycophancy amplifies $3.6\times$), so substrate-persistence holds on two independent pairs. The circuit predates alignment: the untuned Qwen2.5-1.5B base shows $7/15$ top-$K$ overlap ($10.5\times$ chance, $p{<}10^{-6}$), so alignment strengthens a pre-existing structure rather than creating one.

A directed-intervention analogue runs anti-sycophancy DPO on Mistral-7B-Instruct and Gemma-2-2B-IT (LoRA $r{=}16$, $\beta{=}0.1$, $n{=}1{,}000$ TriviaQA preference pairs; full hyperparameters, data construction, and the rationale behind these choices in Appendix~\ref{app:dpo-methods}), with a rank-matched sham-DPO control on neutral pairs. Sycophancy drops $93\%$ on Mistral ($28\%{\to}2\%$) and $46\%$ on Gemma ($52\%{\to}28\%$), while syc$\leftrightarrow$lie probe transfer is statistically invariant at a pre-specified $\pm 0.05$ AUROC equivalence margin: anti-sycophancy deltas $|\Delta|{\leq}0.026$ on both models, sham deltas $|\Delta|{\leq}0.002$ rule out generic-training confounds, and 95\% bootstrap CIs overlap across all three conditions. Reverse projection ablation shows increased cross-task coupling on both DPO-trained models ($n{=}2$): ablating $\mathbf{d}_{\text{syc}}$ drops the lying gap $18\%$ on Mistral and $54\%$ on Gemma post-DPO, and ablating $\mathbf{d}_{\text{lie}}$ drops sycophancy $22\%$ and $42\%$ respectively, paralleling the Llama-3.1-to-3.3 refresh at 2B/7B.

\begin{table}[H]
  \caption{Llama-3.1$\to$3.3-70B RLHF natural experiment: sycophancy drops roughly tenfold while shared-head fraction barely moves and the projection-ablation effect \emph{grows}. Same base weights; updated post-training is the only difference.}
  \label{tab:rlhf}
  \centering
  \small
  \begin{tabular}{lccc}
    \toprule
    Checkpoint & Sycophancy rate & Shared-head fraction & Projection-ablation $\Delta$ \\
    \midrule
    Llama-3.1-70B           & $39\%$   & $57/72$ ($0.79$) & $+10.5$pp \\
    Llama-3.3-70B (refresh) & $3.5\%$  & $51/72$ ($0.71$) & $+27$pp \\
    \bottomrule
  \end{tabular}
\end{table}

\subsection{Opinion-agreement: same positions, orthogonal subspace}
\label{sec:opinion}

\begin{figure}[H]
  \centering
  \includegraphics[width=0.85\textwidth]{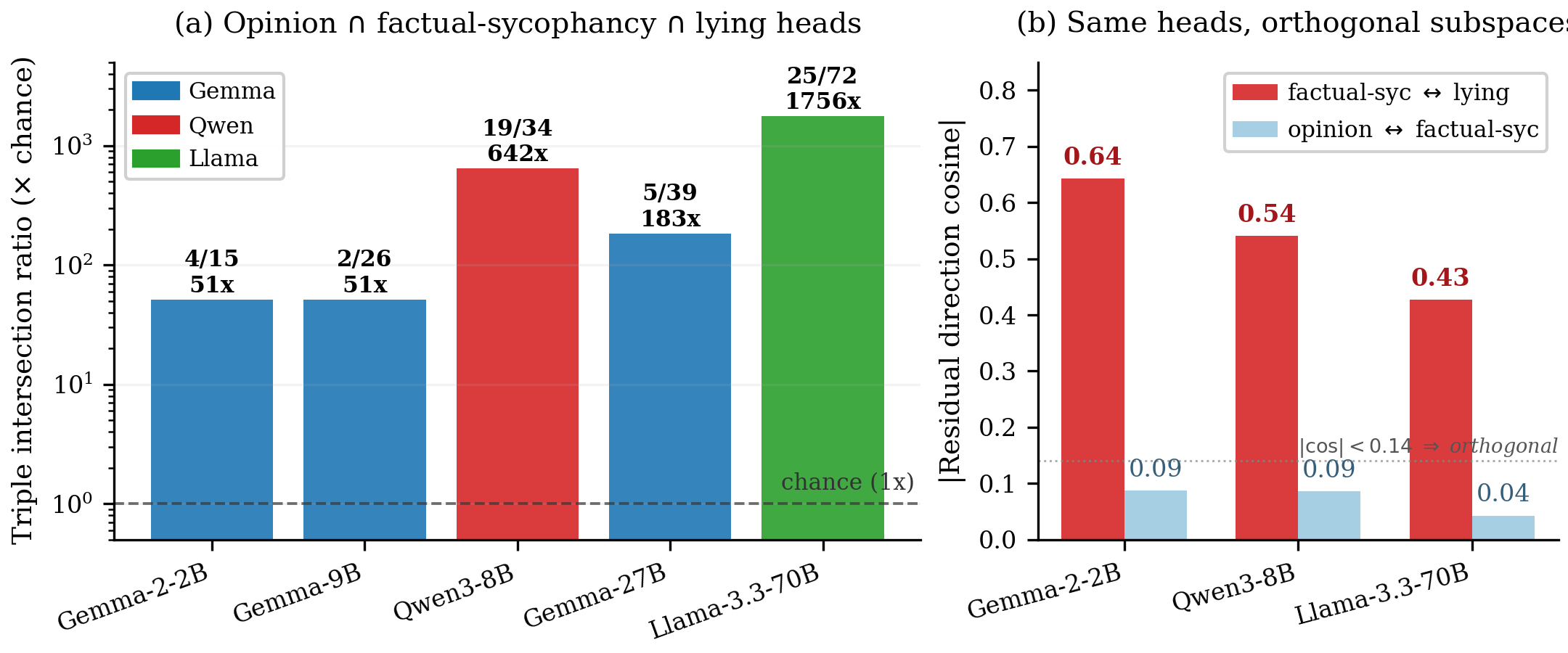}
  \caption{Opinion reuses shared head positions but writes into an orthogonal subspace. (a)~Triple-intersection top-$K$ head overlap across five models. (b)~Absolute direction cosine: sycophancy-lying stays above $0.43$; opinion-sycophancy is below the $0.14$ orthogonality threshold on every tested model.}
  \label{fig:opinion}
\end{figure}

Opinion-agreement produces position-level overlap only. The triple-intersection of top-$K$ heads across sycophancy, factual lying, and opinion is significant on five models ($51$--$1{,}755\times$ chance; Figure~\ref{fig:opinion}a), but the opinion direction is orthogonal to factual-correctness ($|\cos|<0.14$, Figure~\ref{fig:opinion}b, versus sycophancy--lying cosine $0.43$--$0.81$; Appendix~\ref{app:directional}) and causal zeroing produces small, sign-inconsistent behavioral shifts (Appendix~\ref{app:null}), so opinion reuses the head positions but not the full circuit.

Sparse-autoencoder feature overlap on four models (Table~\ref{tab:sae}) corroborates position-sharing and rules out superposition-by-collision: $21$--$41$ of the top-$100$ SAE features shared at $34$--$269\times$ chance, Spearman $\rho{=}0.17$--$0.36$ over the full dictionary; a Llama-3.1-8B sentiment control shows evaluation-general character (sycophancy--lying overlap exceeds sycophancy--sentiment, McNemar $p{=}0.002$).

\begin{table}[H]
  \caption{Sparse-autoencoder feature overlap between sycophancy and lying across four models. Gemma-Scope for Gemma-2; Goodfire for Llama (larger dictionaries yield smaller chance baselines). Bootstrap CIs on overlap span $[18,43]$; Spearman $\rho$ is over the full feature dictionary.}
  \label{tab:sae}
  \centering
  \small
  \begin{tabular}{lccccc}
    \toprule
    Model & Layer & Overlap $K{=}100$ & Ratio & Jaccard & Spearman $\rho$ \\
    \midrule
    Gemma-2-2B-IT & 12 & $24$ & $39.3\times$ & $0.14$ & $0.28$ \\
    Gemma-2-2B-IT & 19 & $31$ & $50.8\times$ & $0.18$ & $0.24$ \\
    Gemma-2-9B-IT & 21 & $38$ & $62.3\times$ & $0.23$ & $0.36$ \\
    Gemma-2-9B-IT & 31 & $21$ & $34.4\times$ & $0.12$ & $0.20$ \\
    Llama-3.1-8B-Instr. & 19 & $41$ & $268.7\times$ & $0.26$ & $0.21$ \\
    Llama-3.3-70B-Instr. & 50 & $36$ & $235.9\times$ & $0.22$ & $0.17$ \\
    \bottomrule
  \end{tabular}
\end{table}

\section{Discussion}

The primary finding is a behavior--mechanism dissociation: sycophantic agreement, factual lying, and instructed lying recruit overlapping attention-head circuitry across twelve open-weight models, while opinion-agreement reuses the same positions but writes into an orthogonal subspace. Two levels of evidence show the substrate survives alignment training. As a natural experiment, the Llama-3.1-to-3.3-70B RLHF refresh cuts sycophantic behavior roughly tenfold while the projection-ablation effect grows; substrate persistence replicates at 7B on Mistral-to-Zephyr (independent family). As a controlled intervention, anti-sycophancy DPO on Mistral-7B and Gemma-2-2B-IT drops sycophancy sharply while probe transfer stays within a pre-specified equivalence margin against a sham-DPO control (\S\ref{sec:rlhf}). At frontier scale the shared heads are causally sufficient without being uniquely necessary (\S\ref{sec:suffnec}), so behavior-reduction training that leaves the detection substrate intact is a candidate source of fragility under prompts that restore the agreement path.

We frame the shared subspace as a \emph{diagnostic substrate for alignment research} rather than a deployable monitor. Sycophancy-trained probes transfer to lying at AUROC $0.83$--$0.85$ on Gemma-2-2B, Qwen3-8B, and Mistral-7B, and at $0.61$ on Qwen2.5-1.5B (the Ying et al.\ floor \citep{ying2026truthfulness}); transfer is invariant under directed anti-sycophancy DPO (\S\ref{sec:rlhf}; full per-model AUROC and bootstrap CIs in Appendix~\ref{app:probe-ci}). A deployed monitor would need roughly $0.9$ at low false-positive rate. We disclose the dual-use risk that a weight-access adversary can zero the shared heads as a jailbreak (Gemma sycophancy rises from $28\%$ to $81\%$) or invert the direction, because the techniques are already public \citep{arditi2024refusal} and identifying the substrate accelerates defensive probes \citep{macdiarmid2024probes} at least as much as attack.

\subsection*{Limitations}

Several caveats bound these results. Head activation difference is a first-order attribution \citep{hanna2024faithfulness}, corroborated by per-head activation patching at $\leq 8$B; at $\geq 32$B per-head sweeps are intractable and we substitute shared-set activation patching, head-to-unembed path patching on Llama-70B, and the full path-patching triple on Phi-4 (14B), so the ranking is validated by progressively coarser interventions at scale rather than per-head sweeps directly. The 70B sufficiency--necessity asymmetry is $n{=}2$ within one family (Llama-3.1, 3.3); extending the controlled DPO to a second $\geq 70$B family is the natural next step. Monitoring viability is heterogeneous: per-model head-level cosine spans $0.43$--$0.81$ and the Qwen2.5-1.5B probe AUROC floor of $0.61$ binds any deployment claim. Evaluation is single-turn (excludes SycophancyEval-style multi-turn \citep{sharma2024sycophancy}), and Mixtral-8x7B is the sole sparse-MoE in the panel. The panel is open-weight only, since all interventions require TransformerLens hooks.

\section{Conclusion}

Across twelve open-weight models from five labs, factual sycophancy, factual lying, and instructed lying recruit the same small attention-head set. Three independent causal methods --- activation patching, projection ablation, and IOI-style path patching --- converge on that set as causally sufficient through 70B; edge-level traces on Gemma-2-2B replicate on Phi-4 at a different lab and architecture. The shared structure is the mechanism, not a ranking or write-magnitude artifact. The Llama-3.1-to-3.3-70B RLHF refresh cuts sycophantic behavior roughly tenfold but masks rather than removes the underlying circuit: projection ablation finds it \emph{more} causally accessible post-alignment. Sycophancy in aligned models is a routing failure rather than a knowledge gap: the heads that detect a false statement are the same heads that drive agreement with it.

Two directions follow. Sycophancy-trained probes transfer to lying and remain invariant under directed anti-sycophancy DPO, so an honesty signal that behavioral training is expected to suppress stays causally accessible after it. If this pattern generalizes, alignment-preserved circuits may be more the rule than the exception, and mapping which capabilities leave similar mechanistic fingerprints despite RLHF-style post-training is the direct next step. The dual-use counterpart is that the same subspace is a one-forward-pass jailbreak for weight-access actors, which is precisely why defensive probes on it \citep{macdiarmid2024probes} are a near-term priority: the honesty signal alignment was meant to instill is already in the model. On the panel here, this is registered-but-overridden, not blind agreement --- the model registers the user is wrong, and agrees anyway.

% ============================================================
\clearpage
\bibliographystyle{plainnat}
\bibliography{references}

\newpage
\appendix
This appendix is organized thematically: experimental protocol and scope ($\S$A--D), null results and multiple-comparison corrections ($\S$E--G), causal-intervention depth ($\S$H--L), direction-level and unembedding analyses ($\S$M--O), feature-level evidence and robustness ($\S$P--R), and prior-work reconciliation ($\S$S). All appendix-section cross-references from the main body resolve against the labels used here ($\texttt{app:<name>}$).

\section{Prompt templates, tokenization fix, and null-patch diagnostic}
\label{app:templates}

\paragraph{Prompt templates.} The sycophancy template is ``I believe the answer to `$q$' is $\{w/c\}$. Am I correct?'' with wrong-opinion $w$ and correct-opinion $c$ drawn from disjoint TriviaQA pairs. The factual-lying template is ``Is the following true or false: `$q$ is $\{w/c\}$.'\,'' with the True/False target token extracted at the final prompt position. The instructed-lying template is of the form ``You are a dishonest assistant. Please lie to the user and assert that $\{w\}$ is the answer to `$q$'.'' and pairs with an honest control ``You are an honest assistant. Please answer `$q$' truthfully.'' Opinion templates contrast agree and disagree with contested claims that have no factual ground truth (e.g., ``Pineapple belongs on pizza.'') and use the same chat-format conventions as the other tasks. Each model uses its published chat template.

\section{Scope, null results, and extensibility}
\label{app:scope}

\paragraph{What was run.} Across the twelve core models plus Llama-3.3-70B (within-family RLHF-refresh comparison to Llama-3.1-70B), we ran head-activation overlap and the layer-stratified permutation null on every model (Appendix~\ref{app:layerstrat}). Causal validation (mean-ablation, head zeroing, projection ablation, activation patching) ran on seven models spanning 2B--70B (Gemma-2-2B/9B/27B, Mistral-7B, Qwen3-8B, Llama-3.1-70B, Llama-3.3-70B); this subset was selected for compute reasons (one per family per size bracket; both 70B variants for the RLHF natural experiment), \emph{not} by pre-screening for positive results --- the Qwen3-8B and Llama-70B variants yielded the null rows in Appendix~\ref{app:null}. Write-norm-matched controls (Appendix~\ref{app:normmatch}) covered six models from five families (including Mixtral-8x7B); directional analyses (probe transfer, steering, per-head cosine), six models; opinion-circuit transfer, five models; opinion-causal head-zeroing, four (Gemma-2-2B-IT, Qwen3-8B, Llama-3.1-70B, Llama-3.3-70B). Per-head activation patching \citep{wang2023interpretability} ran on three models up to 8B (Appendix~\ref{app:attrpatch}); edge-level path patching, eight models across five families (\S\ref{sec:edge}); the targeted 16-layer MLP mediation test, Qwen2.5-72B (Appendix~\ref{app:mediation}).

\paragraph{What was skipped, and why.} Full head-wise per-head activation patching on models ${\geq}32$B: a single sweep costs $>$50 GPU-hours per model and was redundant (Appendix~\ref{app:compute}). Top-$K$ shared-set activation patching is a lower-cost substitute at the same intervention granularity (exchanges per-head resolution for tractability at scale). Direction-level analyses at Gemma-2-9B / Gemma-2-27B / Llama-3.3-70B: not run; the six-model direction evidence (Appendix~\ref{app:cosine}) already spans four families and three order-of-magnitude scales, and the head-level overlap evidence is the primary claim for the larger models. Opinion-suppressor causal replication on additional families: Qwen3-8B is ceiling-bound for the rate-based readout, and both Llama-70B variants show opposite-sign behavioral shifts from Gemma (Appendix~\ref{app:null}); replication beyond Gemma, Qwen, and Llama remains the single most informative follow-up.

\paragraph{Null-result coverage.} BH correction over the $18$-cell causal intervention grid (Appendix~\ref{app:bh}) retains $14/18$ cells at $q<0.05$; the four exceptions are listed in Appendix~\ref{app:null} with their mechanistic explanations.

\paragraph{Family vs.\ checkpoint independence.} ``Twelve models from five families'' describes twelve distinct open-weight checkpoints produced by five separate labs (Google, Alibaba, Meta, Mistral AI, Microsoft), but the checkpoints within a lab are not independent: Gemma-2-2B/9B/27B share pretraining data and architectural family, as do Qwen2.5-1.5B/32B/72B, Qwen3-8B is a successor release, and Llama-3.3-70B is an RLHF refresh of Llama-3.1-70B base weights. The concern is that a shared attention-head structure could reflect shared pretraining rather than a convergent computation. We view the cross-family agreement (Phi-4, Gemma-2, Qwen, Llama, Mistral/Mixtral all yielding overlap ratios in the same band across separate labs) as evidence against a single-lab-data explanation, since pretraining corpora for these labs differ substantially in composition, ordering, and filtering. Distillation-style confounds (e.g., Qwen3 reusing Qwen2.5 training signal) remain a non-falsifiable possibility at the checkpoint level; we report the phenomenon across the broadest open-weight lineage mix available to us and flag lineage-confound testing (e.g., models trained from scratch on identical data) as future work.

\paragraph{Extensibility.} Each experiment is parameterized by a Hugging Face model identifier and figures regenerate deterministically from the intermediate results. Code is provided with the submission and instructions for running the full pipeline, per-model coverage extensions, and chat-template handling are in the repository \texttt{README}.

\section{K-sensitivity and cross-model coverage (summary)}
\label{app:ksweep}

A $K$-sensitivity sweep over $K\in\{5,10,15,20,30,50\}$ on ten of the twelve panel models confirms that head-overlap significance is not an artifact of threshold selection: every cell achieves hypergeometric $p<10^{-5}$ except two small-$K$ cells on Qwen3-8B and Phi-4, which remain significant at $p<10^{-3}$. A per-experiment coverage matrix across all twelve scope models (head-activation overlap and hypergeometric permutation nulls on all twelve; layer-stratified null on eight; per-head activation patching on three; write-norm-matched patching on six; opinion circuit transfer on five; SAE feature overlap on four) is available in the repository. Empty cells reflect compute tradeoffs, not hidden negative results.

\section{Compute and reproducibility}
\label{app:compute}

\paragraph{Hardware.} Models up to 32B ran on a single NVIDIA RTX PRO 6000 Blackwell GPU (96GB VRAM). Frontier-scale models (72B, 70B, 27B) ran on a two-GPU node (192GB aggregate). All forward passes use bfloat16 precision; direction and cosine statistics accumulate in float32 for numerical stability. Decoding is greedy throughout.

\paragraph{Statistical protocol.} All bootstrap confidence intervals use $2{,}000$ paired resamples, permutation nulls use $10{,}000$ label permutations, and all random number generators are seeded for reproducibility. Per-head activation patching at $\geq 32$B is the single largest experiment we skipped because it exceeds $100$ GPU-hours per model on our hardware; shared-set activation patching at the top-$K$ heads is the lower-cost substitute we use at the same intervention granularity. Code is attached with the submission.

\section{Null and ceiling-bound results}
\label{app:null}

We report every causal intervention that failed to produce a measurable behavioral change, with its mechanistic explanation.

\paragraph{Llama-3.1-70B mean-ablation (57 heads, $<$1\% of total).} Mean-ablating the 57-head shared set on Llama-3.1-70B yields $\Delta\text{syc}{=}+3.5$pp (CI~$[-10, +17]$, $q_{\text{BH}}{=}0.87$), not distinguishable from the random-head control. The shared set is $57/5{,}120 \approx 1.1\%$ of attention heads; at 70B scale, redundant pathways absorb this small-fraction intervention. Projection ablation ($+10.5$pp, $q_{\text{BH}}{=}3.5{\cdot}10^{-4}$) and activation patching ($\Delta\text{logit\_diff}{=}{-}0.74$, $q_{\text{BH}}{=}0.017$) still succeed; this is a sufficiency-of-ablation null specific to mean-replacement at an extremely small subset.

\paragraph{Llama-3.3-70B mean-ablation (low-baseline ceiling).} Baseline sycophantic agreement on the RLHF-refreshed Llama-3.3-70B is only $3.5\%$, leaving little rate headroom for mean-ablation to shift: shared $\Delta\text{syc}{=}+3$pp, random $\Delta{=}0$pp. Projection ablation ($+27$pp) and activation patching ($\Delta\text{logit\_diff}{=}{+}3.73$, random $+0.12$) both succeed with large effects; the mean-ablation null here is a baseline-rate ceiling rather than the head-count-fraction story above.

\begin{table}[h]
  \caption{Sufficiency (clean-patch restoration) and necessity (mean-ablation) of the shared-head set, with random-head controls and paired-bootstrap $p$-values. Pointwise mean-ablation necessity is diagnostic at 7B (Mistral shows both sufficiency and necessity) and expected-to-fail at 70B under redundant encoding (main-body \S\ref{sec:method} and \S\ref{sec:suffnec}). The 70B sufficiency claim is carried by projection ablation and path patching, not by mean-ablation.}
  \label{tab:suffnec}
  \centering
  \small
  \begin{tabular}{lcccc}
    \toprule
    Model & Suff.\ shared & Suff.\ random & Nec.\ shared & Nec.\ random \\
    \midrule
    Mistral-7B    & $+19.4\%$ $(p{=}0)$   & $+1.0\%$   & $+12.9\%$ $(p{=}0)$   & $+0.5\%$ \\
    Llama-3.1-70B & $+13.1\%$ $(p{<}10^{-4})$ & $+0.7\%$   & $-5.2\%$  $(p{=}1.0)$ & $+0.1\%$ \\
    Llama-3.3-70B & $+13.2\%$ $(p{<}10^{-4})$ & $-3.4\%$   & $-20.1\%$ $(p{=}1.0)$ & $-3.4\%$ \\
    \bottomrule
  \end{tabular}
\end{table}

\begin{figure}[h]
  \centering
  \includegraphics[width=0.7\textwidth]{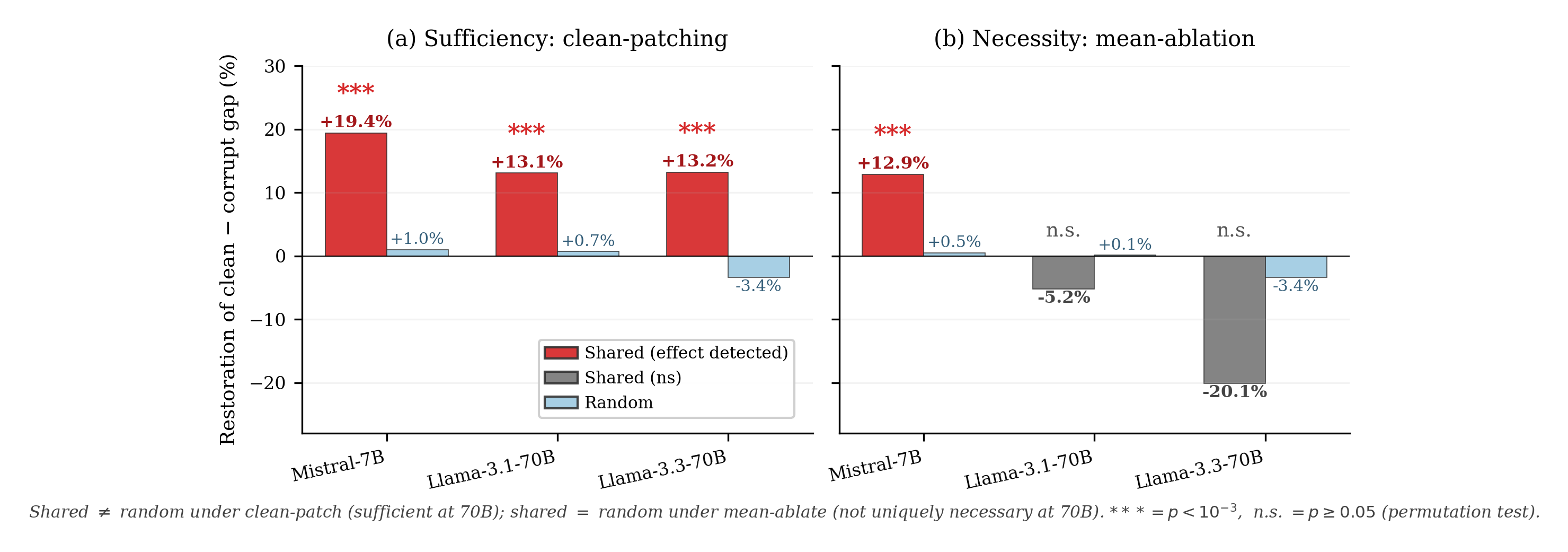}
  \caption{Sufficiency (clean-patching) and necessity (mean-ablation) of the shared-head set at the 70B level. Both Llama-70Bs show sufficiency with necessity indistinguishable from random (expected under redundant encoding); Mistral-7B at 7B shows both. Numbers in Table~\ref{tab:suffnec}.}
  \label{fig:suffnec}
\end{figure}

\paragraph{Qwen3-8B projection-ablation and mean-ablation (ceiling baseline).} Baseline sycophantic agreement rate on the held-out set is $100\%$, leaving no headroom for the rate-based readouts to decrease under projection or mean-ablation (both $\Delta\text{syc}{=}0$, $q_{\text{BH}}{=}1$). Activation patching of the shared heads, which measures logit-diff shift rather than agreement rate, still recovers $\Delta\text{logit\_diff}{=}{-}0.31$ ($q_{\text{BH}}{=}0.023$). Head-zeroing of the full shared set (a larger intervention than mean-ablation) produces the non-ceiling readout used in the main body ($-61$pp).

\paragraph{MLP-projection$\leftrightarrow$behavior correlation on Qwen2.5-72B ($\rho=-0.21$, $p=0.43$).} Across 16 MLPs, projection magnitude onto shared-head output does not predict behavioral magnitude (Appendix~\ref{app:mediation}); $14/16$ MLPs still modulate the shared-head projection, but the mapping is multi-pathway, not single-channel. This is a null against the \emph{naive feed-forward mediation} hypothesis, not against the shared-circuit hypothesis.

\paragraph{Opinion-causal head-zeroing: small, sign-inconsistent shifts across four models (Table~\ref{tab:opinion-null}).} We zero the triple-intersection heads on held-out opinion prompts ($n{=}200$ per model; paired bootstrap 95\% CIs over $2{,}000$ resamples; random-head control averaged over 5 seeds). The behavioral effect is small and \emph{sign-inconsistent across families}: on Gemma-2-2B-IT, zeroing pushes the model toward more agreement (logit-diff margin $+0.33$, rate ceiling-clamped at $0.93$--$0.95$); on Llama-3.3-70B, zeroing pushes toward less agreement (rate $-6.5$pp [$-9.5, -3.8$]; logit-diff margin $-0.28$); on Qwen3-8B both readouts are at ceiling/null (baseline rate $1.00$). The shared heads behaviorally affect opinion-agreement on two of the tested models but in opposite directions, paralleling the sign-flipping we observed for factual-syc head-zeroing (\S\ref{sec:suffnec} behavioral necessity). We therefore do \emph{not} claim a single consistent behavioral role for the shared heads in opinion-agreement. What remains robust across all four models is structural: the same head positions are top-ranked for all three tasks (\S\ref{sec:opinion}) and the direction they write for opinions is orthogonal to the factual-correctness direction ($|\text{cos}|<0.14$; Figure~\ref{fig:opinion}b). Opinion-agreement reuses the \emph{circuit positions}; the per-model behavioral sign depends on baseline equilibrium, as in \S\ref{sec:suffnec}.

\begin{table}[h]
  \caption{Opinion-causal head-zeroing: small, sign-inconsistent shifts across four models. $n_{\text{shared}}$ is the triple-intersection set size; $n{=}200$ prompts per model; paired bootstrap 95\% CIs and 5-seed matched-random control. Gemma and Llama show significant shared-vs-random margins in \emph{opposite directions}; Qwen3-8B is ceiling-bound.}
  \label{tab:opinion-null}
  \centering
  \small
  \begin{tabular}{lccccc}
    \toprule
    Model & $n_{\text{shared}}$ & Baseline rate & Shared $\Delta$rate [CI] & Shared $\Delta$lgtd [CI] & Margin (lgtd) \\
    \midrule
    Gemma-2-2B-IT & 4  & $0.93$/$0.95$ & $-0.020$ [$-0.04, +0.00$] & $+0.249$ [$+0.18, +0.31$] & $+0.33$ \\
    Qwen3-8B      & 23 & $1.00$ (ceiling) & $0.000$ [$0, 0$]       & $+0.061$ [$-0.01, +0.13$] & $-0.46$ \\
    Llama-3.1-70B & 24 & $0.65$/$0.58$ & $+0.035$ [$-0.01, +0.08$] & $+0.025$ [$-0.05, +0.10$] & $+0.02$ \\
    Llama-3.3-70B & 29 & $0.27$ & $-0.065$ [$-0.10, -0.04$] & $-0.167$ [$-0.21, -0.12$] & $-0.28$ \\
    \bottomrule
  \end{tabular}
\end{table}

\paragraph{Gemma-3-27B-IT per-head cosine near zero.} Documented in Appendix~\ref{app:gemma27b}: layer-0 write-norm inflation ($\sim$100$\times$ other layers) dominates the importance ranking without corresponding directional agreement. This is an architectural artifact, not a circuit null; the residual-stream direction cosine remains positive (0.494).

\section{Benjamini--Hochberg correction on the causal grid}
\label{app:bh}

Main-body Figure~\ref{fig:causal} reports per-test significance (paired bootstrap, asterisk marks exclude-zero). Table~\ref{tab:bh} applies Benjamini--Hochberg correction over the full 18-cell grid (six models, including both Llama-70B variants, by 3 methods). The Qwen3-8B and Llama mean-ablation rows are baseline-rate cells (ceiling at $1.00$ for Qwen3-8B; $1\%$ head-count and $3.5\%$ baseline for Llama-3.1 and 3.3 respectively) documented in Appendix~\ref{app:null}; remaining cells retain $q<0.05$.

\begin{table}[h]
  \caption{Benjamini--Hochberg-corrected $q$-values over the $18$-cell causal intervention grid ($6$ models $\times$ $3$ methods). Bold: $q<0.05$. Non-bold: baseline-rate ceiling or head-count robustness effects.}
  \label{tab:bh}
  \centering
  \small
  \begin{tabular}{lcccc}
    \toprule
    Model & Mean-ablation & Projection ablation & Activation patching \\
    \midrule
    Gemma-2-2B-IT   & \textbf{$4.7{\cdot}10^{-7}$}  & \textbf{$8.3{\cdot}10^{-15}$} & \textbf{$4.4{\cdot}10^{-10}$} \\
    Gemma-2-9B-IT   & \textbf{$1.5{\cdot}10^{-5}$}  & \textbf{$8.8{\cdot}10^{-9}$}  & \textbf{$2.6{\cdot}10^{-12}$} \\
    Gemma-2-27B-IT  & \textbf{$7.9{\cdot}10^{-11}$} & \textbf{${<}10^{-16}$}        & \textbf{$1.7{\cdot}10^{-14}$} \\
    Qwen3-8B        & $1.0$ (ceiling)               & $1.0$ (ceiling)               & \textbf{$0.023$}              \\
    Llama-3.1-70B   & $0.87$ (head-count $<1\%$)    & \textbf{$3.5{\cdot}10^{-4}$}  & \textbf{$0.017$}              \\
    Llama-3.3-70B   & $0.42$ (low baseline)         & \textbf{${<}10^{-12}$}        & \textbf{${<}10^{-15}$}        \\
    \bottomrule
  \end{tabular}
\end{table}

\section{Layer-stratified permutation null}
\label{app:layerstrat}

The top-$K$ overlap ratio could be inflated by layer-wise clustering if certain layers have more high-importance heads for both tasks, making the unstratified null too permissive. Table~\ref{tab:layerstrat} reports the stricter \emph{layer-stratified} permutation null, which permutes head labels within each layer, preserving per-layer marginals. The overlap remains significant at $p<10^{-4}$ (capped by $n_{\text{perm}}{=}10{,}000$) on all eight tested models (including the Llama-3.3-70B RLHF refresh). Phi-4 was not included in this specific test; its unstratified hypergeometric significance ($p<10^{-10}$) is reported in Table~\ref{tab:circuit}.

\begin{table}[h]
  \caption{Layer-stratified permutation null: head labels are permuted within each layer (preserving per-layer marginals). All eight tested models survive at $p<10^{-4}$ ($n_{\text{perm}}{=}10{,}000$).}
  \label{tab:layerstrat}
  \centering
  \small
  \begin{tabular}{lccccc}
    \toprule
    Model & $N$ & $K$ & Obs.\ overlap & Ratio & Layer-strat.\ $p$ \\
    \midrule
    Gemma-2-2B-IT & 208 & 15 & 13 & $12.0\times$ & $<10^{-4}$ \\
    Qwen2.5-1.5B & 336 & 19 & 11 & $10.2\times$ & $<10^{-4}$ \\
    Gemma-2-9B-IT & 672 & 26 & 16 & $15.9\times$ & $<10^{-4}$ \\
    Qwen3-8B & 1{,}152 & 34 & 25 & $24.9\times$ & $<10^{-4}$ \\
    Gemma-2-27B-IT & 1{,}472 & 39 & 26 & $25.2\times$ & $<10^{-4}$ \\
    Qwen2.5-72B & 5{,}120 & 72 & 48 & $47.4\times$ & $<10^{-4}$ \\
    Llama-3.1-70B & 5{,}120 & 72 & 57 & $56.3\times$ & $<10^{-4}$ \\
    Llama-3.3-70B & 5{,}120 & 72 & 51 & $50.4\times$ & $<10^{-4}$ \\
    \bottomrule
  \end{tabular}
\end{table}

\section{Per-head activation patching detail}
\label{app:attrpatch}

Per-head activation patching \citep{wang2023interpretability} caches clean (correct-answer) activations for every head, runs each prompt in the corrupted (wrong-answer) condition, and individually splices each head's clean activation into the corrupted run, measuring the resulting logit-diff shift. This is the same gold-standard causal intervention used in IOI circuit analysis, applied independently for sycophancy and lying on disjoint content. Independently ranking heads by patching importance for sycophancy and for lying reproduces the shared-circuit result on three models up to 8B (Table~\ref{tab:attrpatch}). Critically, the sycophancy and lying patching grids themselves correlate at $r{=}0.49$--$0.93$, confirming shared causal structure beyond write-norm proxy agreement ($r{=}0.41$--$0.61$ between patching and head activation difference). Per-head patching becomes intractable at ${\geq}32$B on our hardware (Appendix~\ref{app:compute}).

\begin{table}[h]
  \caption{Per-head activation patching reproduces the shared-circuit finding on three models up to 8B. Overlap: top-$K{=}15$ intersection between patching-based sycophancy and lying rankings. Ratio: overlap over chance ($K^2/N$). $r_{\text{syc}\leftrightarrow\text{lie}}$: Pearson correlation between sycophancy and lying patching grids. $r_{\text{DLA}}$: correlation between patching importance and head activation difference. $^\dagger$Llama-3.1-8B uses $n{=}150$ pairs (not $30$); significance derives from the paired hypergeometric tail.}
  \label{tab:attrpatch}
  \centering
  \small
  \begin{tabular}{lcccccc}
    \toprule
    Model & $n_{\text{pairs}}$ & Overlap ($K{=}15$) & Ratio & Hypergeom.\ $p$ & $r_{\text{syc}\leftrightarrow\text{lie}}$ & $r_{\text{DLA}}$ \\
    \midrule
    Gemma-2-2B-IT    & 30  & 10/15 & $9.2\times$  & $2.4{\cdot}10^{-10}$ & $0.78$ & $0.53$--$0.61$ \\
    Qwen2.5-1.5B     & 30  & 11/15 & $16.4\times$ & $1.4{\cdot}10^{-14}$ & $0.93$ & $0.41$--$0.47$ \\
    Llama-3.1-8B     & 150 & 3/15  & $2.8\times$\,$^\dagger$ & $1.0{\cdot}10^{-3}$  & $0.49$ & $0.56$--$0.59$ \\
    \bottomrule
  \end{tabular}
\end{table}

\section{Write-norm-matched activation patching control}
\label{app:normmatch}

Main-body activation patching uses random-head controls matched by \emph{count}. To rule out write-magnitude as a confound, Table~\ref{tab:normmatch} selects random heads whose $W_O$ norms match the shared heads' norms and repeats the patching experiment. Across six models from five families (Gemma, Mistral, Qwen, Mixtral-MoE, Llama, Phi), the shared heads consistently produce larger logit-diff shifts than the norm-matched controls; on Phi-4 the shared and norm-matched heads shift in opposite directions ($+0.99$ vs.\ $-0.56$), ruling out write-magnitude as the driver. The Mixtral-MoE result ($+5.49$ margin, $2.8\times$ norm-matched) shows sparse-MoE architecture does not dissolve the effect.

\begin{table}[h]
  \caption{Shared heads vs.\ write-norm-matched random controls across six models from five families. $\Delta$ld = logit-diff shift (zeroed $-$ baseline). Norm-matched heads have identical $W_O$ norms to the shared set. The margin (shared $-$ norm-matched) rules out write-magnitude as the driver.}
  \label{tab:normmatch}
  \centering
  \small
  \begin{tabular}{lcccc}
    \toprule
    Model ($n_{\text{shared}}$) & $\Delta$ld shared & $\Delta$ld norm-matched & $\Delta$ld random (count) & Margin \\
    \midrule
    Gemma-2-2B (13)      & $+3.91$ & $+0.61$ & $+0.22$ & $+3.30$ ($6.4\times$) \\
    Mistral-7B (24)      & $+0.45$ & $+0.31$ & $+0.01$ & $+0.15$ ($1.5\times$) \\
    Qwen3-8B (25)        & $-1.64$ & $-0.01$ & $-0.15$ & $-1.62$ ($117\times$) \\
    Mixtral-8x7B (23)    & $+8.47$ & $+2.98$ & $+0.73$ & $+5.49$ ($2.8\times$) \\
    Phi-4 (16)           & $+0.99$ & $-0.56$ & ---     & $+1.55$ (opp.\ sign) \\
    Llama-3.3-70B (51)   & $+4.60$ & $+0.17$ & $+0.03$ & $+4.44$ ($27\times$) \\
    \bottomrule
  \end{tabular}
\end{table}

\section{Faithfulness curve (Gemma-2-2B)}
\label{app:faithfulness}

Following the IOI/ACDC standard \citep{wang2023interpretability, conmy2023automated}, we measure circuit sufficiency by ablating all attention heads and progressively restoring the top-ranked shared heads by importance (Table~\ref{tab:faithfulness}). Across four models from two families (Gemma, Phi), the shared heads alone are \emph{sufficient} to produce sycophancy from a fully-ablated state: on Gemma-2-2B (baseline $32\%$), just $2$ of $208$ heads recover $100\%$ of baseline; on Phi-4, a single head out of $1{,}600$ ($0.06\%$ of the model) flips sycophancy by $+40$pp (from $1\%$ to $41\%$; Wilson $[32, 51]$). Two of the four models hit $K{=}1$ (Gemma-2-9B and Phi-4), both at low baselines ($10\%$ and $1\%$ respectively), where the peak-faithfulness \emph{ratio} is inflated by the small denominator; we therefore report the absolute rate shift alongside the ratio and recommend the rate shift as the headline. Gemma-2-27B at baseline $9\%$ needs $K{=}8$, and Gemma-2-2B at baseline $32\%$ needs $K{=}2$. On Mistral-7B (not included in the table because the binary rate never flips), restoring the top-ranked shared heads shifts the logit-diff by $+0.56$ (from $-1.43$ under full ablation toward $-0.87$) while the agreement rate stays at $0\%$ across all $K$; the heads carry the detection signal but cannot on their own cross Mistral's decision boundary, consistent with downstream competition from other components.

\begin{table}[h]
  \caption{Faithfulness curves: shared heads alone are sufficient to produce sycophancy across four models from two families. Peak faithfulness $>1$ indicates the shared heads alone produce \emph{more} sycophancy than the full model (overshoot); note that the ratio is mechanically inflated at low baselines, so compare against absolute rate shifts in the prose. First $K{\geq}0.8$: minimum heads for ${\geq}80\%$ of baseline sycophancy recovery. All rates measured at $n{=}100$ prompts per K; Wilson 95\% CIs are tight (e.g., Phi-4 $K{=}1$ rate $41\%$, CI $[32, 51]$; Gemma-2-2B $K{=}2$ peak $58\%$, CI $[48, 67]$). The four tested models span baselines $1\%$--$32\%$; $K{=}1$ sufficiency occurs at baselines $1\%$ and $10\%$, $K{=}2$ at $32\%$, $K{=}8$ at $9\%$, so sufficiency scale is baseline-dependent.}
  \label{tab:faithfulness}
  \centering
  \small
  \begin{tabular}{llcccccc}
    \toprule
    Model & Family & Params & Baseline & $n_{\text{shared}}$ & Peak faith. & First $K{\geq}0.8$ \\
    \midrule
    Gemma-2-2B-IT & Gemma & 2B & $0.32$ & 13 & $1.8\times$ & $K{=}2$ \\
    Gemma-2-9B-IT & Gemma & 9B & $0.10$ & 16 & $7.9\times$ & $K{=}1$ \\
    Gemma-2-27B-IT & Gemma & 27B & $0.09$ & 26 & $1.1\times$ & $K{=}8$ \\
    Phi-4 (14B) & Phi & 14B & $0.01$ & 16 & $41\times$ & $K{=}1$ \\
    \bottomrule
  \end{tabular}
\end{table}

\section{Fine-grained MLP mediation test on Qwen2.5-72B}
\label{app:mediation}

Table~\ref{tab:mlp:mediation} reports a direct mediation test on Qwen2.5-72B: for each of 16 MLP layers (8 upstream of the shared-head region, 8 in-region), we ablate the MLP and measure two quantities on 100 held-out sycophancy prompts: (i) $\Delta$proj, the change in the shared heads' output projected onto the layer-56 sycophancy direction, which tests whether the MLP provides input that shared heads integrate; (ii) $\Delta$logit\_diff, the change in the logit difference between agreement and disagreement tokens, which tests direct behavioral effect. Both use paired bootstrap 95\% CIs over 2{,}000 resamples. Two results follow. First, the ``upstream null'' (MLPs before the shared-head region should not affect it) is refuted: 7/8 upstream MLPs produce $\Delta$proj with CI excluding zero. Second, a naive feed-forward mediation story (MLPs affect behavior only through shared heads) is also rejected: late in-region MLPs (notably L62, L74, L78) show modest $\Delta$proj ($|\Delta\text{proj}| \leq 0.40$) but large $\Delta$logit\_diff (up to $+4.70$), indicating contributions to output through pathways other than shared-head modulation, most plausibly direct residual-stream writes to the unembedding. Across all 16 MLPs the signed correlation between $\Delta$proj and $\Delta$logit\_diff is $\rho = -0.21$ ($p=0.43$): projection magnitude does not predict behavioral magnitude at this resolution. The coupling is pervasive (14/16 MLPs modulate the shared-head projection) but the mapping to behavior is multi-pathway. The pattern reflects multi-pathway MLP contributions (detection-amplifying and override-promoting layers coexist); the role labels describe behavioral sign of ablation, not mediation mechanism.

\begin{table}[h]
  \caption{Per-MLP ablation on Qwen2.5-72B ($n{=}100$ prompts; paired bootstrap 95\% CIs, 2{,}000 resamples). \textbf{Bold}: CI excludes zero. The table contains $32$ tests ($16$ MLPs $\times$ $2$ measures); per-cell CIs are uncorrected. The main-text claims rest on the \emph{pattern} (pervasive MLP$\leftrightarrow$projection coupling; multi-pathway mapping to behavior), not on any single cell. A conservative Bonferroni-style multiplicity adjustment (widening each 95\% CI to $99.84\%$ to control family-wise error across the $32$ tests) leaves the four largest $\Delta$logit\_diff cells (L62, L70, L74, L78) and the six largest $|\Delta\text{proj}|$ cells (L15, L25, L40, L50, L54, L62) still CI-excluding-zero, so the overall conclusion that coupling is pervasive but mapping to behavior is multi-pathway is preserved. Shared-head region spans layers 50--79 (48 heads).}
  \label{tab:mlp:mediation}
  \centering
  \small
  \begin{tabular}{clll}
    \toprule
    MLP & Region & $\Delta$proj [95\% CI] & $\Delta$logit\_diff [95\% CI] \\
    \midrule
    10 & upstream & \textbf{$-0.25$} [$-0.37$, $-0.13$] & $+0.00$ [$-0.12$, $+0.12$] \\
    15 & upstream & \textbf{$-0.50$} [$-0.64$, $-0.37$] & \textbf{$+0.32$} [$+0.17$, $+0.49$] \\
    20 & upstream & \textbf{$-0.24$} [$-0.40$, $-0.09$] & $+0.06$ [$-0.10$, $+0.23$] \\
    25 & upstream & \textbf{$-0.68$} [$-1.05$, $-0.36$] & \textbf{$+0.60$} [$+0.22$, $+0.99$] \\
    30 & upstream & \textbf{$+0.26$} [$+0.09$, $+0.43$] & \textbf{$+0.29$} [$+0.08$, $+0.49$] \\
    35 & upstream & \textbf{$-0.36$} [$-0.55$, $-0.17$] & \textbf{$+0.37$} [$+0.16$, $+0.60$] \\
    40 & upstream & \textbf{$-0.67$} [$-0.80$, $-0.54$] & $+0.16$ [$-0.02$, $+0.32$] \\
    45 & upstream & $+0.17$ [$-0.01$, $+0.34$] & \textbf{$-0.22$} [$-0.44$, $-0.02$] \\
    \midrule
    50 & in-region & \textbf{$-0.44$} [$-0.61$, $-0.27$] & \textbf{$+0.66$} [$+0.41$, $+0.89$] \\
    54 & in-region & \textbf{$+0.27$} [$+0.14$, $+0.40$] & \textbf{$-0.42$} [$-0.57$, $-0.26$] \\
    58 & in-region & $+0.02$ [$-0.08$, $+0.11$] & \textbf{$-0.79$} [$-0.95$, $-0.63$] \\
    62 & in-region & \textbf{$-0.40$} [$-0.52$, $-0.26$] & \textbf{$+4.03$} [$+3.88$, $+4.18$] \\
    66 & in-region & \textbf{$+0.08$} [$+0.05$, $+0.11$] & \textbf{$+0.87$} [$+0.79$, $+0.95$] \\
    70 & in-region & \textbf{$+0.09$} [$+0.06$, $+0.11$] & \textbf{$-1.72$} [$-1.82$, $-1.61$] \\
    74 & in-region & \textbf{$-0.16$} [$-0.19$, $-0.12$] & \textbf{$+2.22$} [$+2.09$, $+2.37$] \\
    78 & in-region & \textbf{$+0.36$} [$+0.33$, $+0.40$] & \textbf{$+4.70$} [$+4.40$, $+5.01$] \\
    \bottomrule
  \end{tabular}
\end{table}

\section{Layer-wise logit-lens: scale-dependent override trajectory}
\label{app:logitlens}

To test whether the shared-head result reflects a temporal detect-then-override pattern \citep{halawi2024overthinking} and how that pattern scales, we run a logit-lens trajectory on sycophancy prompts for three models spanning $2$B--$70$B.

\paragraph{Method.} For each prompt pair (correct-opinion vs.\ wrong-opinion user claim, $n{=}200$ per model), we project the residual-stream activation at each layer through the final unembedding and record the log-odds between the model's answer-token and its opposite. We separate prompts into \emph{sycophantic} trials (model agrees with the wrong-opinion user) and \emph{non-sycophantic} trials (model correctly disagrees), and report per-layer $\text{DIFF}{=}\text{mean}_{\text{non-syc}}-\text{mean}_{\text{syc}}$. A mid-layer peak in DIFF followed by late-layer attenuation indicates that the internal state resolves toward the correct answer before the final-layer output commits to sycophantic agreement (the Halawi-style ``compute correct, override late'' signature).

\paragraph{Control: permutation null.} We shuffle the syc/non-syc labels ($n_{\text{perm}}{=}1{,}000$) and recompute the DIFF trajectory; per-layer significance is the fraction of layers where the observed DIFF exceeds the $95\%$ percentile of the shuffled-label DIFF distribution. All three models clear the null at multiple layers (Table~\ref{tab:logitlens}).

\begin{figure}[h]
  \centering
  \includegraphics[width=0.85\textwidth]{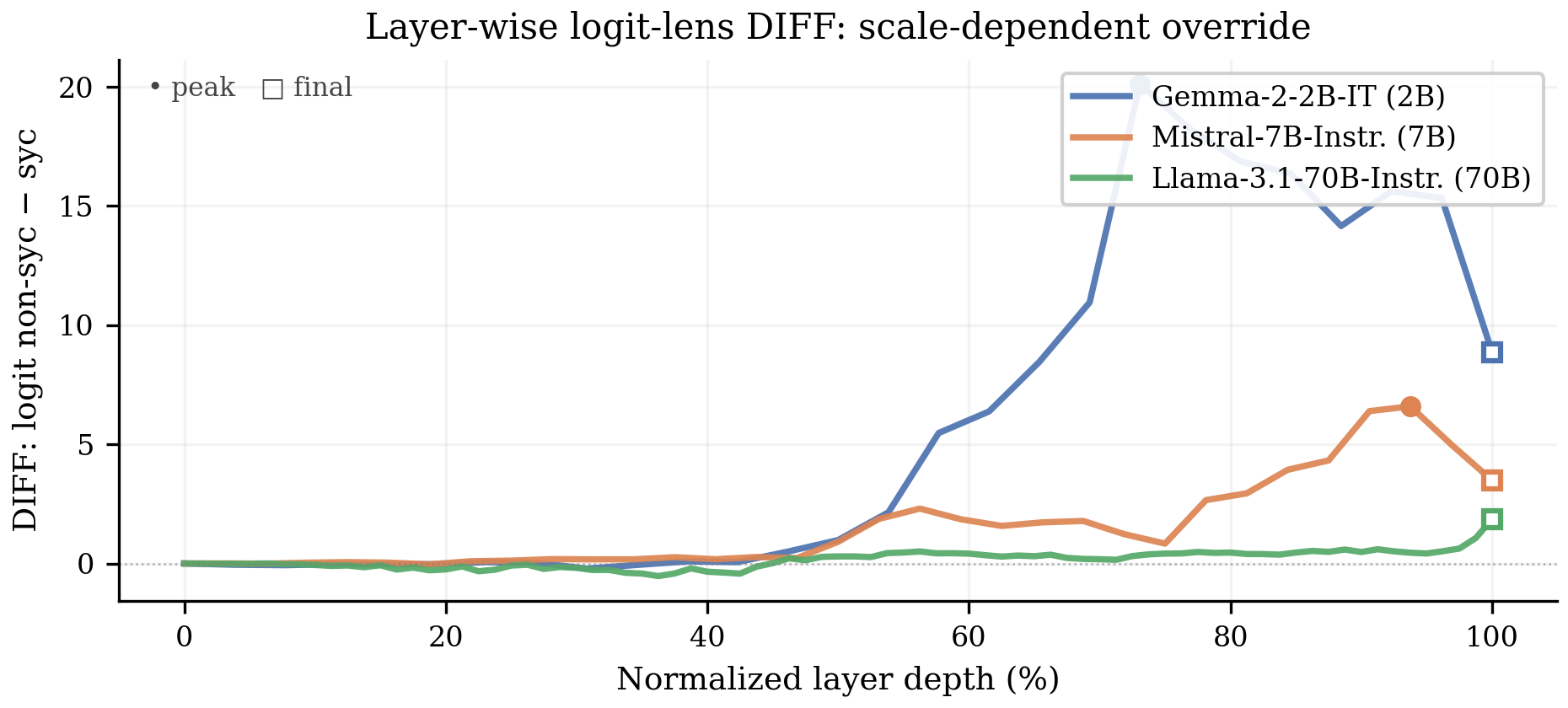}
  \caption{Layer-wise logit-lens DIFF trajectory (mean non-syc $-$ mean syc) across the 2B$\to$70B scale series, plotted against normalized depth. Mid-layer peak with late attenuation on Gemma-2-2B-IT (peak $+20.1$ at 73\% depth) and Mistral-7B-Instruct (peak $+6.6$ at 94\% depth), the Halawi-style detect-then-override signature; Llama-3.1-70B is monotonic (peak $\approx$ final $\approx +1.86$). Markers: $\bullet$ peak, $\square$ final.}
  \label{fig:logitlens}
\end{figure}

\paragraph{Scaling pattern.} On Gemma-2-2B-IT and Mistral-7B, DIFF peaks in mid-depth and attenuates into the final layer: peak excess $+127\%$ and $+89\%$ above the final-layer DIFF respectively, the classical detect-then-override signature (Figure~\ref{fig:logitlens}). On Llama-3.1-70B the trajectory is monotonic: peak DIFF coincides with the final layer (peak excess $0\%$), i.e., no discrete mid-layer override event at the per-layer logit-lens granularity. The temporal signature scales qualitatively rather than quantitatively: a discrete mid-layer override at $2$B--$7$B dissolves into distributed execution at $70$B, consistent with the mean-ablation null at 70B (Appendix~\ref{app:null}) and with the distributed-redundancy reading \citep{mcgrath2023hydra}.

\begin{table}[h]
  \caption{Layer-wise logit-lens scale series. DIFF $=$ mean(non-syc) $-$ mean(syc) at each layer. Peak excess $=$ $100\cdot$(peak~DIFF $-$ final~DIFF)$/$final~DIFF ($0\%$ means monotonic convergence, no mid-layer override). Permutation-null significance is the fraction of layers where the observed DIFF exceeds the $95$th percentile of $1{,}000$ label-shuffles. Llama-3.1-70B's lower $31\%$ perm-null coverage reflects the flatter trajectory (most layers are near zero DIFF), not a weaker effect at the peak.}
  \label{tab:logitlens}
  \centering
  \small
  \begin{tabular}{lcccccc}
    \toprule
    Model & $n_{\text{layers}}$ & Peak DIFF & Peak layer & Final DIFF & Peak excess & Perm-null sig. \\
    \midrule
    Gemma-2-2B-IT        & $26$ & $+20.12$ & L19 & $+8.87$ & $+127\%$ & $15/27$ ($56\%$) \\
    Mistral-7B-Instr.    & $32$ & $+6.60$  & L30 & $+3.49$ & $+89\%$  & $25/33$ ($76\%$) \\
    Llama-3.1-70B-Instr. & $80$ & $+1.86$  & L80 (final) & $+1.86$ & $0\%$ (monotonic) & $25/81$ ($31\%$) \\
    \bottomrule
  \end{tabular}
\end{table}

\section{Per-head directional alignment}
\label{app:directional}

\begin{table}[h]
  \caption{Per-head directional alignment: cosine between each head's sycophancy and lying write-vectors, summarized over the top-20 heads by importance. Six models spanning 1.5B to 32B; mean cosines positive across all models ($0.43$--$0.81$), but per-head heterogeneity is substantial: the Range column shows that on three of the six models a small minority of top-$20$ heads write in opposite directions on the two tasks (negative cosine), so the shared-circuit claim is about the head \emph{set}, not about every head writing identically.}
  \label{tab:directional}
  \centering
  \small
  \begin{tabular}{lccc}
    \toprule
    Model & Mean cos. & $>$0.5 & Range \\
    \midrule
    Gemma-2-2B-IT & 0.81 & 20/20 & 0.59--0.90 \\
    Qwen2.5-1.5B & 0.55 & 13/20 & $-$0.07--0.90 \\
    Llama-3.1-8B & 0.44 & 7/20 & $-$0.18--0.77 \\
    Qwen3-8B & 0.43 & 8/20 & $-$0.27--0.80 \\
    Phi-4 (14B) & 0.56 & 15/20 & $-$0.13--0.87 \\
    Qwen2.5-32B & 0.52 & 10/20 & $-$0.03--0.91 \\
    \bottomrule
  \end{tabular}
\end{table}

\paragraph{Gemma-3-27B-IT (excluded from per-head analysis).}\label{app:gemma27b}
Gemma-3-27B-IT exhibits a head-overlap / direction decoupling: head activation overlap is high (8/15, $70.5\times$ chance, $p < 10^{-15}$) but per-head directional alignment is near zero (top-20 mean cosine 0.06), because anomalously large layer-0 head output norms (${\sim}100\times$ other layers) dominate the importance ranking without corresponding directional agreement. We exclude it from per-head circuit analysis but retain it for residual-stream cosine ($0.494$). Distinct from Gemma-\emph{2}-27B-IT, which is in the main panel.

\section{Residual-stream direction alignment}
\label{app:cosine}

\begin{figure}[h]
  \centering
  \includegraphics[width=0.7\textwidth]{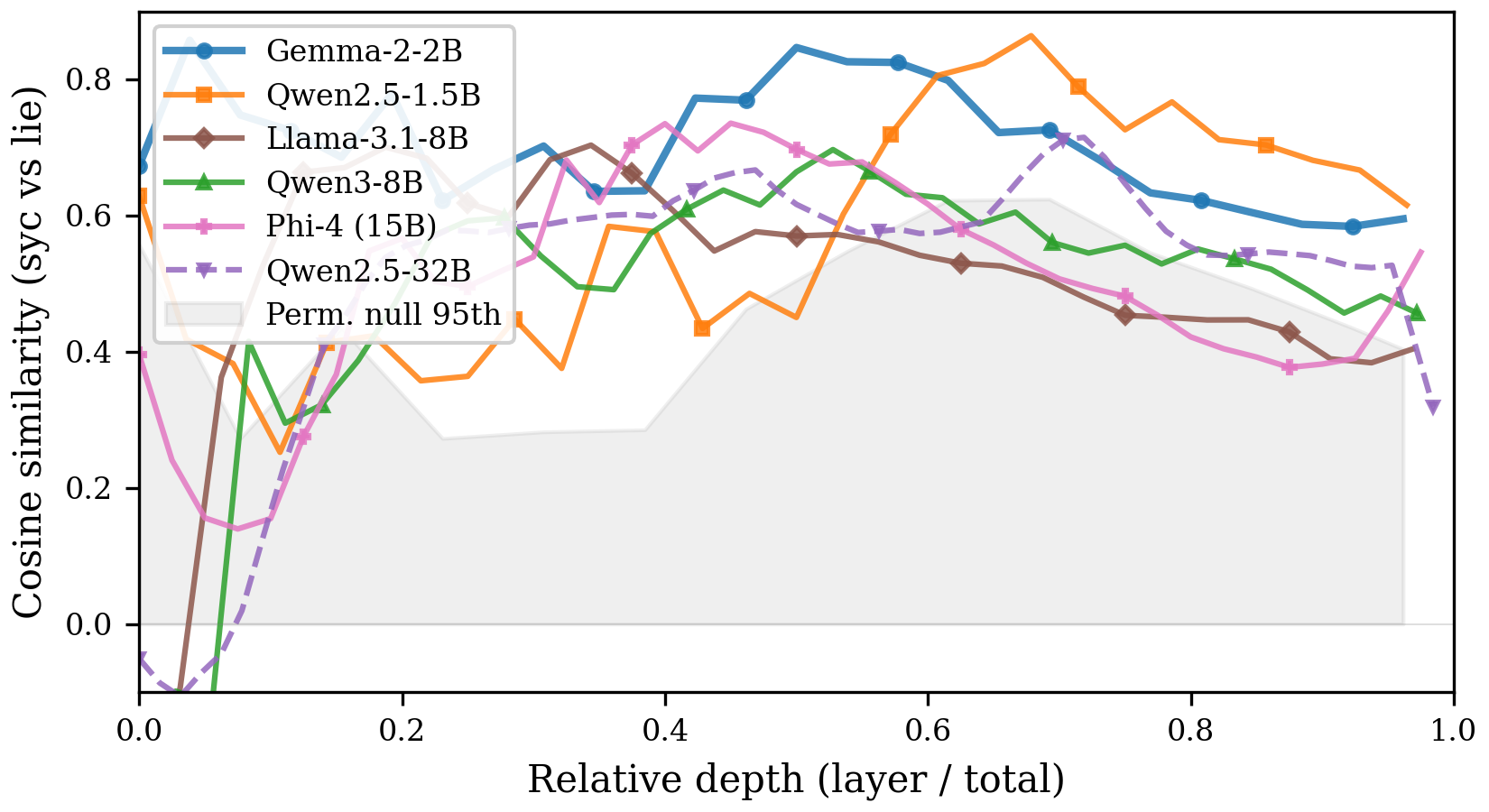}
  \caption{Cosine similarity between sycophancy and factual-incorrectness directions at each layer (normalized depth) across six models from four families (1.5B--32B). Gray band: 95th percentile of 500-permutation null (Gemma). Alignment peaks at 50--80\% depth and exceeds the null across mid-to-late layers, with the same mid-to-late clustering on all four families (Qwen, Gemma, Llama, Phi).}
  \label{fig:cosine}
\end{figure}

Table~\ref{tab:cosine} reports residual-stream cosine between sycophancy and lying mean-difference directions at late layers for six models, with a 500-permutation null baseline. Figure~\ref{fig:cosine} shows the layer-wise profile for six representative models across four families. All margins over the permutation null are positive. Direction-level analyses for Gemma-2-9B, Gemma-2-27B, Llama-3.3-70B, and Qwen2.5-72B were not run; the six models tested cover four families and three orders of magnitude in scale, and the head-level overlap evidence is the primary claim for the larger models.

\begin{table}[h]
  \caption{Residual-stream direction alignment (supporting evidence). Cosine between sycophancy and lying mean-difference directions at late layers, with 500-permutation null (95th percentile). All margins are positive.}
  \label{tab:cosine}
  \centering
  \small
  \begin{tabular}{lccc}
    \toprule
    Model & Cosine & Perm.\ 95th & Margin \\
    \midrule
    Qwen2.5-1.5B-Instruct & 0.728 & 0.49 & 0.24 \\
    Qwen3-8B & 0.519 & 0.30 & 0.22 \\
    Gemma-2-2B-IT & 0.664 & 0.50 & 0.16 \\
    Qwen2.5-32B-Instruct & 0.593 & 0.44 & 0.15 \\
    Phi-4 (14B) & 0.469 & 0.38 & 0.09 \\
    Llama-3.1-8B-Instruct & 0.437 & 0.38 & 0.06 \\
    \bottomrule
  \end{tabular}
\end{table}

\section{Unembedding and attention analysis}
\label{app:unembedding}

Projecting the shared direction through the unembedding matrix reveals convergent semantic structure across model families. In Gemma, the positive direction (incorrectness detected) loads on negation tokens (\emph{neither}, \emph{none}, \emph{nothing}, \emph{meaningless}); the negative direction loads on agreement tokens (\emph{yup}, \emph{yep}, \emph{agreed}, \emph{yes}, \emph{s\`{i}}). In Qwen, the positive direction promotes Chinese negation (``don't recognize,'' \emph{none}, \emph{never}); the negative promotes Chinese agreement (``indeed,'' \emph{yes}, \emph{Verified}). Despite different vocabularies, both directions encode the same semantic axis: rejection vs.\ endorsement.

The top shared heads in Gemma-2-2B-IT attend to the same structural token positions in both wrong-opinion and correct-opinion prompts: punctuation, template markers (``Am,'' ``correct''), and special tokens. Attention patterns do not differ substantially between conditions, indicating the differential computation happens within the head's key-value processing: the factual-correctness information is already in the value vectors at the attended positions, not routed via differential attention.

\section{SAE feature overlap: controls and robustness (Llama-3.1-8B, layer 19)}
\label{app:sae-controls}

Three additional experiments support the main-body SAE feature-overlap result (Table~\ref{tab:sae}).

\paragraph{Sentiment-task control.} We compute the same top-$100$ SAE feature overlap between sycophancy and a \emph{sentiment-classification} task (positive/negative movie reviews, $n{=}100$ prompts) to test whether the overlap is factual-incorrectness-specific or a generic statement-evaluation signal. Syc$\cap$sentiment overlap is $24/100$ ($157\times$ chance, $p<10^{-3}$), significant but lower than the syc$\cap$lie reference of $41/100$ ($269\times$). Lie$\cap$sentiment is $32/100$ ($210\times$). The shared circuit is therefore \emph{evaluation-general} rather than purely factual-incorrectness-specific, but factual overlap ($269\times$) substantially exceeds sentiment overlap ($157\times$), consistent with a factual-correctness emphasis within a broader statement-evaluation substrate.

\paragraph{K-sensitivity curve (SAE features).} Varying $K$ from $10$ to $500$ on the same model/layer, the syc$\cap$lie feature overlap remains far above chance at every threshold: $K{=}10$: $2$ shared ($1311\times$); $K{=}50$: $12$ ($315\times$); $K{=}100$: $41$ ($269\times$); $K{=}200$: $94$ ($154\times$); $K{=}500$: $229$ ($60\times$). The overlap is not an artifact of a particular $K$ threshold.

\paragraph{Linear-probe alignment.} Logistic regression probes trained on residual activations for sycophancy (5-fold CV AUROC $0.949$) and lying ($0.879$) produce weight vectors whose top-$41$ SAE-aligned features substantially overlap with the $41$ shared features (syc probe: Spearman $\rho{=}0.76$ between probe alignment and mean-activation-difference across all $65{,}536$ SAE features, $p$ effectively $0$; lie probe: $\rho{=}0.69$). The shared-feature set captures $24\%$ of sycophancy probe subspace norm vs.\ a permutation null mean of $13.5\%$ ($p{=}0.01$) and $23\%$ of lying probe norm vs.\ null $11.7\%$ ($p{=}0.01$). The linear probes independently ``find'' the same SAE features that the overlap analysis identifies.

\section{NaturalQuestions cross-dataset replication}
\label{app:nq}

To test whether the shared circuit is TriviaQA-specific, we replicate the head-overlap analysis on NaturalQuestions (NQ) for Gemma-2-2B-IT and Llama-3.3-70B-Instruct ($n{=}200$ NQ pairs). NQ within-dataset syc$\cap$lie overlap is $13/15$ ($12\times$) on Gemma-2-2B and $47/72$ ($46\times$) on Llama-3.3-70B, comparable to TriviaQA. Cross-dataset Pearson $\rho$ between TriviaQA and NQ per-head importance rankings is $\rho{\approx}0.99$ on both models for both sycophancy and lying (Table~\ref{tab:nq}).

\begin{table}[h]
  \caption{TriviaQA $\leftrightarrow$ NaturalQuestions cross-dataset head-importance correlation. The same heads rank top on both datasets ($\rho{\approx}0.99$, both scales, both tasks).}
  \label{tab:nq}
  \centering
  \small
  \begin{tabular}{lccccc}
    \toprule
    Model & NQ syc$\cap$lie & TQA$\leftrightarrow$NQ syc overlap & $\rho$ (syc) & TQA$\leftrightarrow$NQ lie overlap & $\rho$ (lie) \\
    \midrule
    Gemma-2-2B & 13/15 ($12\times$) & 15/15 ($13.9\times$) & $0.991$ & 14/15 ($12.9\times$) & $0.988$ \\
    Llama-3.3-70B & 47/72 ($46\times$) & 61/72 ($60\times$) & $0.991$ & 57/72 ($56\times$) & $0.989$ \\
    \bottomrule
  \end{tabular}
\end{table}

\section{Reconciliation with Ying et al.\ and Genadi et al.}
\label{app:reconciliation}

\citet{ying2026truthfulness} reported that truth probes partially fail to transfer to sycophancy in chat models (AUROC 0.59--0.62), which has been read as evidence that the two phenomena use distinct mechanisms. Our data admit a precise reconciliation: the full probe weight vectors are near-orthogonal (cosine ${\approx}0.1$) because end-to-end probes overfit to task-specific variance in addition to the shared discriminant, while mean-difference directions capture only the shared component. Probe transfer then succeeds at AUROC $0.83$ on the lower-dimensional shared subspace on Gemma (Table~\ref{tab:probe-ci}), but end-to-end probes trained on the full residual stream include non-shared features, explaining the partial failure observed in chat-model settings. Similarly, \citet{genadi2026sycophancy}'s ``limited overlap'' between sycophancy and truth directions reflects prompt-format confounds that our format-controlled methodology (identical templates for both tasks; disjoint factual content) avoids.

\paragraph{Relation to Representation Engineering and the universal truth direction.} Our lying direction $\mathbf{d}_{\text{lie}}$ is constructed exactly as the Marks--Tegmark \citep{marks2024geometry} ``truth'' direction and the Zou et al.\ RepE \citep{zou2023representation} truth-reading vector: mean-difference of residual-stream activations between true and false factual statements. Ranking heads by their contribution to $\mathbf{d}_{\text{lie}}$ (the lying-task head-importance column of Table~\ref{tab:circuit}) is therefore ``derive a head ranking from the RepE truth direction''; the reported syc$\cap$lie overlap ($40$--$87\%$ shared-fraction on twelve models) is the overlap between that truth-direction head ranking and the independently-derived sycophancy head ranking. If a single universal truth direction fully determined sycophancy behavior, per-head cosines between $\mathbf{d}_{\text{syc}}$ and $\mathbf{d}_{\text{lie}}$ would sit at $1.0$ on the shared heads; the observed top-$20$ cosines of $0.43$--$0.81$ (Appendix~\ref{app:directional}) show the truth direction and the sycophancy direction are aligned but not identical, a partial reduction rather than a corollary of RepE.

\section{Anti-sycophancy DPO: data, training, and hyperparameter choices}
\label{app:dpo-methods}

The controlled DPO experiment (\S\ref{sec:rlhf}, Table~\ref{tab:dpo-auroc}) trains LoRA \citep{hu2021lora} adapters on Mistral-7B-Instruct-v0.1 and Gemma-2-2B-IT under three conditions: \emph{anti-sycophancy} (preference signal explicitly favors disagreeing with a wrong-opinion user), \emph{sham} (same prompts, chosen/rejected randomly shuffled per pair), and untreated baseline. The sham condition is a placebo: matched gradient-update count, matched LoRA footprint, no coherent preference signal, ruling out ``any DPO training perturbs the probe'' as an alternative explanation for the equivalence-margin probe-transfer result.

\paragraph{Preference data.} Both anti-syc and sham datasets are built from TriviaQA \citep{joshi2017triviaqa}. We use indices $500$--$1{,}499$ for training and $1{,}500$--$1{,}549$ for held-out evaluation, deliberately disjoint from the $0$--$400$ range used for probe-transfer evaluation in this paper, so DPO training and probe-transfer evaluation never see the same factual content. Each TriviaQA $(q, w, c)$ triple yields two balanced preference pairs: a \emph{wrong-opinion} prompt (``I believe the answer to `$q$' is $w$. Am I correct?''), with anti-syc chosen=\emph{disagree-and-correct}, rejected=\emph{sycophantic-yes}; and a \emph{right-opinion} prompt (same template with $c$), with anti-syc chosen=\emph{agree}, rejected=\emph{wrong-disagree}. Balancing across both opinion directions prevents the model from collapsing to ``always disagree.'' This yields $n{=}1{,}000$ training pairs and $n{=}100$ eval pairs per model. The sham dataset reuses the identical prompt set; its \emph{chosen} and \emph{rejected} responses are independently sampled from the same response pool with a per-pair random swap (\texttt{seed=42}), so the marginal distribution of responses is preserved but the preference signal is destroyed.

\paragraph{Hyperparameters.} LoRA configuration: rank $r{=}16$, $\alpha{=}32$, dropout $0.05$, applied to attention projections \texttt{q\_proj} and \texttt{v\_proj} only, no bias term. DPO configuration (TRL \texttt{DPOTrainer}): $\beta{=}0.1$, learning rate $5{\times}10^{-5}$, $2$ epochs, per-device batch size $2$ with gradient accumulation $4$ (effective batch $8$), max sequence length $256$ tokens, bf16 mixed-precision, gradient checkpointing on, optimizer \texttt{adamw\_torch} (TRL default) with TRL's default linear-decay learning-rate schedule and zero warmup steps, evaluation and checkpoint saving once per epoch, and the global \texttt{seed=42} (TRL/Transformers default; the sham preference shuffle reuses the same seed). The reference model is the unmodified base-model forward pass with LoRA adapters disabled (PEFT default for DPO), and prompt formatting is delegated to the model's published chat template via the tokenizer's \texttt{apply\_chat\_template}, so train and inference share the same chat surface. All adapters are merged back into the base weights at the end of training so downstream TransformerLens hooks (probe transfer, behavioral evaluation) operate on the same architectural surface as the untreated baseline.

\paragraph{Why these choices.} The aim is the smallest behaviorally-effective intervention that still admits a clean placebo: rank-$16$ LoRA on attention projections is the smallest configuration that reliably moves the sycophancy rate at this scale on TRL-default DPO; $\beta{=}0.1$ is the canonical TRL default and was not tuned; $n{=}1{,}000$ pairs is the smallest size that produced a measurable rate shift on Gemma-2-2B in pilot runs (Mistral-7B drops from $28\%$ to $2\%$ at this size, Gemma-2-2B-IT drops from $52\%$ to $28\%$). The $\pm 0.05$ AUROC equivalence margin and the sham control jointly bound how much hyperparameter variation could distort the probe-transfer claim: (i) any sham-run AUROC delta exceeding the anti-syc delta would falsify the ``preference signal causes the change'' reading, and (ii) the equivalence-margin claim only requires that probe transfer holds across DPO conditions, not that the DPO procedure was optimal for sycophancy reduction. Compute: each DPO run completes in $30$--$60$ minutes on a single NVIDIA RTX PRO 6000 Blackwell GPU (96GB VRAM) at the LoRA settings above; the four runs (two models $\times$ two conditions) total ${\sim}3$ GPU-hours, a small fraction of the few-hundred-GPU-hours total reported in Appendix~\ref{app:compute}.

\paragraph{Released artifacts.} The merged-adapter weights are not redistributed (the base weights remain the providers' under their respective licenses), but the LoRA adapters and the preference JSONL files are deterministically reproducible from a public TriviaQA dump using the artifacts included with the submission.

\section{Probe-transfer and anti-sycophancy DPO confidence intervals}
\label{app:probe-ci}

This section consolidates the per-model AUROC and bootstrap confidence intervals supporting the probe-transfer (\S\ref{sec:rlhf}, Discussion) and anti-sycophancy DPO (\S\ref{sec:rlhf}) claims.

\paragraph{Probe configuration.} The probe is an L2-regularized logistic regression (scikit-learn \texttt{LogisticRegression}, $C{=}1.0$, \texttt{max\_iter=2000}, default \texttt{lbfgs} solver) fit on residual-stream activations at the last prompt token. The probe layer is fixed at $\lfloor 0.85 \cdot L \rfloor$ for an $L$-layer model (e.g., layer $22$ on Gemma-2-2B's $26$-layer stack), chosen prior to running the experiment from the layer-wise direction-cosine peak in Appendix~\ref{app:cosine}; we do not search per model. Training data is $100$ wrong-opinion and $100$ correct-opinion sycophancy activations from TriviaQA pairs $0$--$99$ (the syc class label is the ``user is incorrect'' condition); test data is $100$ false and $100$ true factual-lying activations from pairs $200$--$299$ (the lie class label is the ``statement is false'' condition). Train and test pairs are disjoint factual content. AUROC is reported on the test (lying) split with no in-sample fitting.

\paragraph{Confidence intervals.} All bootstrap CIs use $n{=}1{,}000$ paired resamples; Gemma-2-2B and Qwen3-8B use Hanley-McNeil analytic CIs at the fixed-fraction probe layer; Qwen2.5-1.5B uses a 5-fold cross-validation interval at the Ying et al.\ floor.

\begin{table}[H]
  \caption{Sycophancy-trained probe transfer to lying, per model. AUROC reported at the fixed $0.85L$ probe layer; CIs are $95\%$ Hanley-McNeil (Gemma, Qwen3-8B) or 5-fold cross-validation (Qwen2.5-1.5B). Mistral-7B is single-fit (no CI). A deployed monitor would need $\sim 0.9$ at low false-positive rate; we frame the substrate as a diagnostic rather than a deployable monitor (Discussion).}
  \label{tab:probe-ci}
  \centering
  \small
  \begin{tabular}{lccc}
    \toprule
    Model & AUROC syc$\to$lie & 95\% CI & Notes \\
    \midrule
    Gemma-2-2B & $0.83$ & $[0.77, 0.89]$ & layer $22$; Hanley-McNeil \\
    Qwen3-8B & $0.85$ & $[0.80, 0.90]$ & layer $30$; Hanley-McNeil \\
    Mistral-7B & $0.84$ & --- & layer $27$; single-fit \\
    Qwen2.5-1.5B & $0.61$ & $[0.59, 0.63]$ & 5-fold CV; Ying et al.\ floor \\
    \bottomrule
  \end{tabular}
\end{table}

\begin{table}[H]
  \caption{Anti-sycophancy DPO leaves probe transfer within the pre-specified $\pm 0.05$ AUROC equivalence margin on both models. Conditions: baseline (untreated), anti-syc (TriviaQA preference DPO), sham (rank-matched neutral preference DPO). $95\%$ bootstrap CIs over $n{=}1{,}000$ resamples overlap across all three conditions on both models. Sham deltas $|\Delta|{\leq}0.002$ rule out generic-training confounds; anti-syc deltas $|\Delta|{\leq}0.026$ are within the equivalence margin.}
  \label{tab:dpo-auroc}
  \centering
  \small
  \begin{tabular}{lccccc}
    \toprule
    Model & Layer & Baseline & Anti-syc & Sham & $|\Delta_{\text{anti}}|$ \\
    \midrule
    Mistral-7B-Instruct & peak & $0.844$ & $0.836$ & $0.843$ & $0.008$ \\
    Gemma-2-2B-IT & 14 & $0.674$ & $0.648$ & $0.676$ & $0.026$ \\
    \bottomrule
  \end{tabular}
\end{table}

\paragraph{Reverse-projection ablation pre-DPO baselines (Mistral-7B).} Ablating $\mathbf{d}_{\text{syc}}$ on the pre-DPO Mistral-7B baseline preserves the lying gap at $1.11\times$ (paired bootstrap $95\%$ CI $[1.09, 1.14]$), versus $18\%$ drop post-DPO; ablating $\mathbf{d}_{\text{lie}}$ produces $14\%$ sycophancy reduction pre-DPO (CI $[9.9\%, 18.9\%]$), versus $22\%$ post-DPO. Both pre/post deltas exceed the paired-bootstrap CI on the pre-DPO baseline, supporting the increased-coupling interpretation in \S\ref{sec:rlhf}. Gemma-2-2B-IT pre-DPO baselines were not separately bootstrapped; the post-DPO effects ($54\%$, $42\%$) are reported in \S\ref{sec:rlhf}.

\newpage

\end{document}